\documentclass[journal]{IEEEtran}

\usepackage{stackengine}
\usepackage{amsmath,amssymb,amsfonts}
\usepackage[ruled,vlined]{algorithm2e}
\usepackage[dvipsnames]{xcolor}
\usepackage{colortbl}
\usepackage{pgfplots}
\usepackage{verbatim}
\pgfplotsset{width=7cm,compat=1.8}
\usepgfplotslibrary{fillbetween}
\usepackage{xcolor}
\usepackage{tikz}
\usepackage{tikzscale}
\usepackage{pgfplots}
\usetikzlibrary{patterns}
\pgfplotsset{every tick/.style={black,}}
\usepackage{subcaption}
\usepackage{caption}
\usepackage{amsmath}
\usepackage{multirow}
\usepackage{multicol}
\usepackage{url}
\usepackage{pgfplotstable}
\usepackage{array}
\usepgfplotslibrary{groupplots}
\usetikzlibrary{patterns}
\usepackage[normalem]{ulem}
\newcommand{\corr}{\mathbf{\rho}}
\usepackage[super]{nth}
\usepackage{tabularx}
\usepackage{bbding}
\usepackage{booktabs}

\newcommand{\CFO}{\text{CFO}}
\newcommand{\PD}{\text{PD}}

\pgfplotsset{compat=1.11,
        /pgfplots/ybar legend/.style={
        /pgfplots/legend image code/.code={%
        \draw[##1,/tikz/.cd,bar width=8pt,yshift=-0.2em,bar shift=0pt]
                plot coordinates {(0cm,0.8em)};},
},
}

\newcommand\xrowht[2][0]{\addstackgap[.5\dimexpr#2\relax]{\vphantom{#1}}}

\newcolumntype{b}{>{\columncolor{blue!10}}c}
\newcolumntype{y}{>{\columncolor{yellow!10}}c}
\newcolumntype{d}{>{\columncolor{red!7}}c}

\newcolumntype{C}{>{\centering\arraybackslash}X}
\newcolumntype{D}{>{\hsize=\dimexpr2\hsize+2\tabcolsep+\arrayrulewidth\relax}C}

\ifCLASSOPTIONcompsoc
  \usepackage[nocompress]{cite}
\else
  \usepackage{cite}
\fi

\begin{document}

\title{PRONTO: \underline{Pr}eamble \underline{O}verhead Reduction with \underline{N}eural Ne\underline{t}works for C\underline{o}arse Synchronization}

\author{
\IEEEauthorblockN{
Nasim Soltani, Debashri Roy, and Kaushik Chowdhury
}\\
\IEEEauthorblockA{
Electrical and Computer Engineering Department, Northeastern University, Boston, MA
}\\
\IEEEauthorblockA{\{soltani.n, d.roy\}@northeastern.edu, krc@ece.neu.edu}}

\markboth{Accepted in IEEE Transactions on Wireless Communications (TWC) March 2023}%
{}

\maketitle

\begin{abstract}
In IEEE 802.11 WiFi-based waveforms, the receiver performs coarse time and frequency synchronization using the first field of the preamble known as the legacy short training field (L-STF). The L-STF occupies upto 40\% of the preamble length and takes upto 32 $\mu$s of airtime. With the goal of reducing communication overhead, we propose a modified waveform, where the preamble length is reduced by eliminating the L-STF. To decode this modified waveform, we propose a neural network (NN)-based scheme called PRONTO that performs coarse time and frequency estimations using other preamble fields, specifically the legacy long training field (L-LTF). Our contributions are threefold: (i) We present PRONTO featuring customized convolutional neural networks (CNNs) for packet detection and coarse carrier frequency offset (CFO) estimation, along with data augmentation steps for robust training. (ii) We propose a generalized decision flow that makes PRONTO compatible with legacy waveforms that include the standard L-STF. (iii) We validate the outcomes on an over-the-air WiFi dataset from a testbed of software defined radios (SDRs). Our evaluations show that PRONTO can perform packet detection with 100\% accuracy, and coarse CFO estimation with errors as small as 3\%. We demonstrate that PRONTO provides upto 40\% preamble length reduction with no bit error rate (BER) degradation. We further show that PRONTO is able to achieve the same performance in new environments without the need to re-train the CNNs.
Finally, we experimentally show the speedup achieved by PRONTO through GPU parallelization over the corresponding CPU-only implementations.
\end{abstract}
 


\section{Introduction}
\label{sec:intro}

The ever-increasing demand for wireless spectrum has led to transformative breakthroughs in the use of massive channel bandwidths (e.g., 2 GHz channels in the 57-72 GHz band)~\cite{massive-bandwidth}, massive multiple input multiple output (MIMO) (with several dozens of antenna elements)~\cite{massive-mimo,mimo-overview}, and physical layer signal processing innovations (e.g., channel aggregation~\cite{channel-aggregation}). All of these methods involve assumptions of availability of specialized hardware, often at the cutting edge of systems design. As opposed to these, in this paper, we propose an approach called PRONTO that reduces the occupancy time of the wireless channel in WiFi networks by shortening packet preambles. Since preambles are present in \emph{every} transmitted packet, even a modest reduction results in a significant cumulative benefit over time. PRONTO is carefully designed to be backwards compatible with legacy WiFi waveforms for general purpose adoption and coexistence with standards-compliant access points (APs) deployed today. 

\begin{figure}[t!!!]
    \centering
    \includegraphics[width=0.45\textwidth]{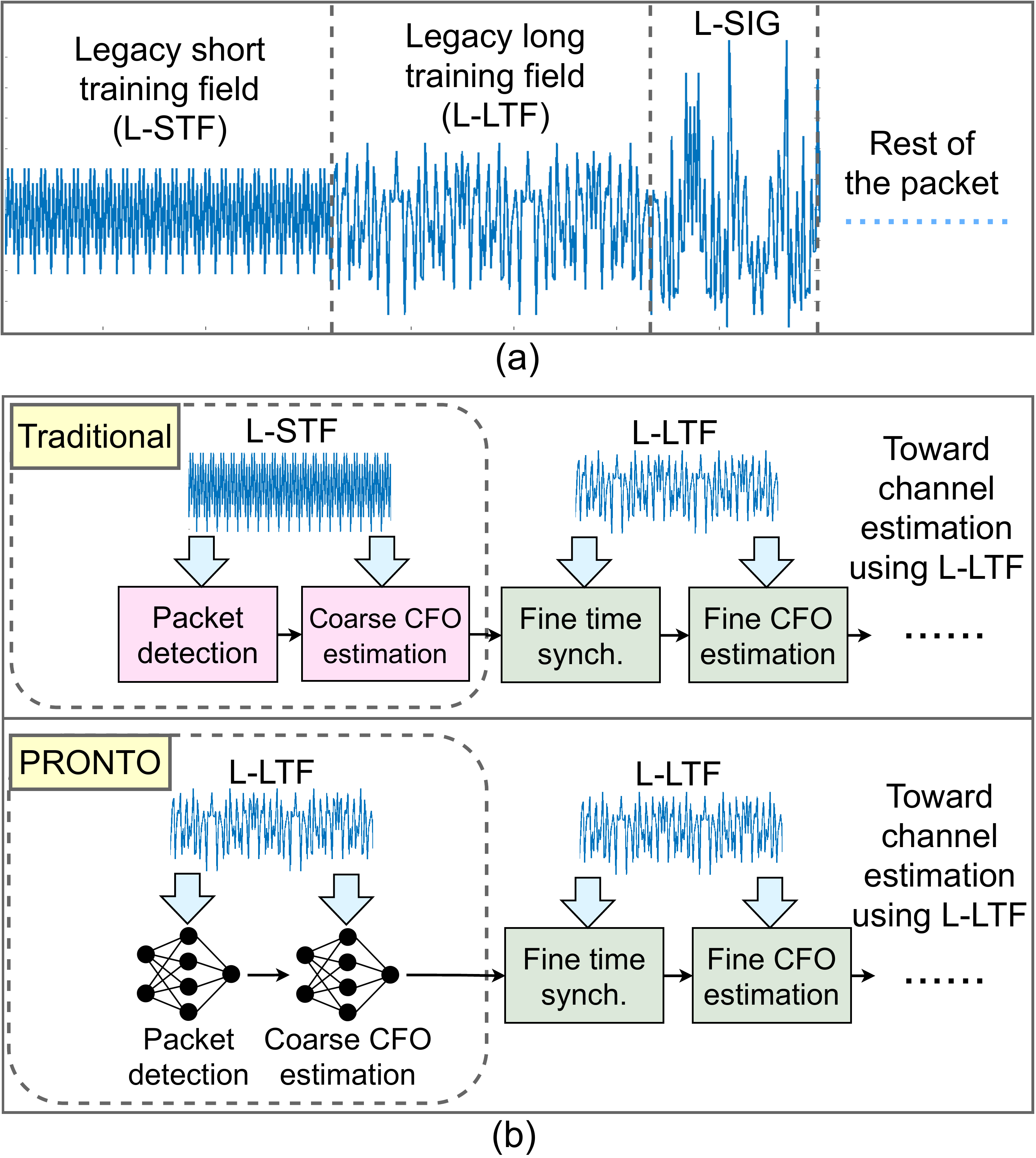}
    \caption{(a) IEEE 802.11 OFDM-family preamble structure. (b) Traditional packet detection and coarse CFO estimation with L-STF, and PRONTO packet detection and coarse CFO estimation without L-STF.\vspace{-0.5cm}}
    \label{fig:overview}
\end{figure}

\noindent $\bullet$ \textbf{Challenges in shortening the preamble:} A typical WiFi packet is composed of a multi-field preamble and data payload. The preamble is used in many tasks, including establishing synchronization between the transmitter and the receiver by removing relative time/frequency offsets. As shown in Fig.~\ref{fig:overview}(a), the preamble in orthogonal frequency division multiplexing (OFDM) family of IEEE 802.11 waveforms starts with the legacy short training field (L-STF) followed by the legacy long training field (L-LTF) that have bits arranged in pre-set configurations with short and long periods, respectively.
First, the L-STF is used to detect the start of the packet (i.e., coarse time synchronization) and then for coarse carrier frequency offset (CFO) estimation. Following this, the L-LTF is used for fine time synchronization and fine CFO estimation. The total CFO is the sum of coarse and fine CFOs, which is later compensated on the whole received packet. Traditional time and frequency synchronization steps are shown in Fig.~\ref{fig:overview}(b). PRONTO revisits the fundamental structure of the WiFi packet by answering the following question: \emph{Can the L-STF be completely eliminated from the preamble, without any adverse impact on decoding?}

\noindent $\bullet$ \textbf{PRONTO approach using deep learning:} In the absence of part of the preamble, the L-STF, PRONTO must extract relevant information from the remaining fields. Towards this goal, it leverages deep learning using convolutional neural networks (CNNs), which have already been demonstrated to be remarkably capable in physical layer signal classification problems~\cite{salvo-classification} such as modulation recognition~\cite{modulation-recognition,deepsig-modulation-ota,zheleva-modulation,zheleva2,modulation-infocom,nasim-dyspan} and RF fingerprinting~\cite{RF-fingerprinting-lora,lora2-infocom,nasim-da,nasim-uav,tong-edge}. In all these problems, CNNs recognize patterns embedded within sequences of in-phase and quadrature (I/Q) samples. Classical methods that perform packet detection and coarse CFO estimation rely on known specific preamble patterns of fixed periods. As opposed to this, once trained, CNNs are able to detect a pattern that is present \emph{somewhere} 
within their input sequences, regardless of its short or long period, and associate each input with a category (classification) or value (regression). In PRONTO, we leverage this ability by training custom-designed CNNs using L-LTF signals (that follow the now eliminated L-STF). \emph{Our hypothesis here is that PRONTO should be able to predict both the start of packet and the coarse CFO, with only the L-LTF as input to the trained CNNs} (Fig.~\ref{fig:overview}(b)).

\noindent $\bullet$ \textbf{PRONTO distinction from previous work:} Reducing preamble length in WiFi systems is a real challenge studied before by the related literature. \cite{liang2007preamble} reduces preamble length by dropping a few training symbols and using a complete iterative receiver algorithm. They achieve preamble  length reduction at the expense of less accurate channel estimation and higher bit error rate (BER). \cite{wang2007preamble} designs a new shorter WiFi preamble by superimposing different fields in the legacy preamble. However, their method is not compatible with conventional WiFi receivers. PRONTO shifts the burden of packet detection and coarse CFO estimation to CNNs without imposing BER degradation, while the rest of the receiver chain remains the same. 
We note that our work is distinct from others that use neural networks as an alternative to the legacy processing blocks, but retain the same inputs and outputs as the latter. Applications of such uses of neural networks exist for channel estimation~\cite{mauro-channelEst,channel-estimation-deep,channel-online,channel2}, signal demapping~\cite{demapper,demapper2}, and decoding~\cite{decoder,decoding2}. 
Among these works are~\cite{packet-detect-cfo},~\cite{packet-detection}, and~\cite{coarse-cfo} that detect packets and estimate CFOs using L-STF signals, similar to the legacy processing blocks.
PRONTO is different from~\cite{coarse-cfo} that models CFO estimation using L-STF signals as a classification problem, where one of the pre-defined classes is chosen as the CFO category instead of predicting an exact CFO value. PRONTO is distinct from~\cite{packet-detect-cfo,packet-detection}, where packet detection using L-STF signals is modeled as a regression problem that yields upto $\sim$8\% false alarm rate.
In terms of overall objectives, our work comes close to prior research on new waveforms, such as removing pilots in orthogonal frequency division multiplexing (OFDM) symbols~\cite{Pilotless}. However, approaches like~\cite{Pilotless} fundamentally change the receiver processing flow. 
As opposed to this, PRONTO is also designed with the goal of maintaining backward compatibility with legacy waveforms. Through a decision logic included alongside PRONTO, the process-flow seamlessly switches between full preamble-enabled packets (i.e., with L-STF) or PRONTO-capable waveforms (i.e., L-STF missing).

We summarize our contributions as follows:
\begin{itemize}
    \item We propose a modified version of the IEEE 802.11 OFDM waveforms where the preamble length is reduced by eliminating the first field (i.e., L-STF). To decode the modified waveform, we propose a deep-learning-based scheme called PRONTO that uses CNNs to detect the packet and estimate coarse CFO using L-LTF. We parameterize our method to be adaptable to signals with different bandwidths, and further propose a generalized decision flow to make the proposed approach compatible with standard waveforms.
    \item The first objective of PRONTO is to perform packet detection as a classification task that returns the start index of the packet, if a part of L-LTF signal exists in the input. We propose a robust training method by artificially shifting L-LTF signal and injecting periods of noise in the training set through a data augmentation step. The CNN trained on this large variety of inputs can detect L-LTF patterns with 100\% accuracy in different noise levels.
    \item The second objective of PRONTO is to perform coarse CFO estimation, for which we train a regressor CNN to return a coarse CFO value for each detected L-LTF. We propose another data augmentation step to dynamically augment the training set by imposing different CFO values to each training signal. The CNN trained on this large variety of training data, is able to predict coarse CFO values in L-LTFs with errors as small as 3\%.
    \item We validate PRONTO on two over-the-air (OTA) WiFi datasets with multiple SNR levels. We demonstrate that PRONTO causes no degradation in BER compared to the traditional WiFi receiver.
    \item We provide computation complexity for PRONTO compared to traditional (i.e., \texttt{MATLAB}-based) algorithms for packet detection and coarse CFO estimation, as well as run-time on server and edge platforms. We show that PRONTO packet detection and coarse CFO estimation (if properly parallelized on GPUs) can perform upto 2053x and 1.92x faster computation, respectively, compared to their traditional \texttt{MATLAB}-based counterparts.
\end{itemize}

The rest of the paper is organized as follows. Section~\ref{sec:traditional} describes the legacy methods for packet detection and coarse CFO estimation using L-STF. Section~\ref{sec:proposed} explains PRONTO scheme in details, as well as the decision flow for backward compatibility. Section~\ref{sec:dataset} describes the datasets and the deep learning setup used to evaluate PRONTO system. Section~\ref{sec:eval} presents the results and Section~\ref{sec:conclusion} concludes the paper.
\section{Background and Preliminaries}
\label{sec:traditional}
In IEEE 802.11 OFDM waveforms, the standard preamble starts with the L-STF. L-STF contains 10 repetitions of a training symbol with length $\eta$. Each training symbol contains one period of a specific pattern, and its length $\eta$ directly depends on the FFT length $\mathcal{N}$, as $\eta=\frac{1}{4}\mathcal{N}$~\cite{ieee-book}. Consequently, the length of full L-STF is $10\eta=\frac{10}{4}\mathcal{N}$. L-STF is modulated with binary phase shift keying (BPSK) modulation. It is not scrambled and has no channel encoding. L-STF is traditionally used for packet detection and coarse CFO estimation, due to its good correlation properties.

\subsection{Packet Detection with L-STF}\label{sec:traditional-packet-detect}
For detecting the packet, the classical method needs the full L-STF signal and a threshold $\beta$, which is a number in the range [0,1], typically close to 1. The classical method finds the start index of the packet by performing an auto-correlation between a portion of L-STF referred to as $\zeta=\text{L-STF}\,(0:10\eta-\eta-1)$, and its delayed version  $\zeta_{\text{delayed}}=\text{L-STF}\,(\eta:10\eta-1)$, where $\eta=\frac{1}{4}\mathcal{N}$.

Next, the complex correlation, $\corr^{(\tau)}$, between $\zeta$ and $\zeta_{\text{delayed}}$ in time $\tau$ is computed as (\ref{eq:rho}), where * is the notation for complex conjugate.
\begin{equation}\label{eq:rho}
 \corr^{(\tau)} = \sum_{i=0}^{\eta-1}\zeta(i+\tau) \times \zeta^*_{\text{delayed}}(i+\tau+\eta)
\end{equation}

The energy vector $E^{(\tau)}$ in the correlation window is calculated as in (\ref{eq:energy}).
\begin{equation}\label{eq:energy}
    E^{(\tau)} = \sum_{i=0}^{\eta-1} \left| \zeta_{\text{delayed}}(i+\tau+\eta) \right|^2
\end{equation}
Then, $\corr^{(\tau)}$ is normalized through dividing it by $E^{(\tau)}$ as in (\ref{eq:rho_norm}).
\begin{equation}\label{eq:rho_norm}
\corr_{\text{norm.}}^{(\tau)} = \frac{|\corr^{(\tau)}|^2}{E^{(\tau)} \times E^{(\tau)}}    
\end{equation}

In the normalized correlation vector, $\corr_{\text{norm.}}^{(\tau)}$, we record the indices of the elements that are greater than $\beta$ as vector $I$, and the number of elements (peaks) as $n_{\corr}$. According to the 802.11 standard, the first peak shows the start of the packet. However, \texttt{MATLAB} implementation performs an additional step in the function \texttt{wlanPacketDetect}~\cite{wlanPacketDetect} based on~\cite{terry2002ofdm}, to ensure the detected pattern is actually L-STF. In this step: First, the number of detected peaks are checked to be larger than or equal to 1.5 times the training symbol length ($n_{\corr} \geq 1.5\times \eta$). Second, the sum of index distances of peaks from the first peak should be smaller than 3 times the symbol length ($\sum_{i=1}^{\eta}(I_{i}-I_{0})<3\times\eta$). These criteria ensure that sufficient number of correlation peaks are detected and they are not too distant from each other. In this case, $I_0 - 1$ is returned as the coarse start time index of the packet.

\subsection{Coarse CFO Estimation with L-STF}\label{sec:traditional-cfo}

The coarse CFO is derived from the phase of complex conjugate of a portion of the L-STF, multiplied by a second portion of the L-STF~\cite{terry2002ofdm}. The time-domain method for coarse CFO estimation, as implemented in \texttt{MATLAB} function \texttt{wlanCoarseCFOEstimate}~\cite{wlanCoarseCFOEstimate}, needs 9 short training symbols for coarse CFO estimation. These 9 symbols are chosen with a distance of 1 symbol from each other. The L-STF (consisting of 10$\times\eta$ short training symbols) starts with a default guard interval of $\mathcal{G}=\frac{3}{4}\eta$, and these samples are considered as offset from the beginning of the signal. After $\mathcal{G}$, two portions of the L-STF, $\chi=\text{L-STF}\,(\mathcal{G}+(0:8\times \eta-1))$ and $\xi=\text{L-STF}\,(\mathcal{G}+(0:8\times \eta-1)+\eta)$ each with length of 8 training symbols are separated and composed.

Next, coarse CFO is calculated using the phase of multiplication of complex conjugate of $\chi$ multiplied by $\xi$ as shown in (\ref{eq:phase}), where $\phi(x)=\arctan\,(\frac{\mathrm{Imag}(x)}{\mathrm{Real}(x)})$ and $f_s$ is sampling frequency in Hz.
\begin{equation}\label{eq:phase}
    \CFO_{\text{coarse}} = \frac{\phi(\chi^*\times \, \xi)\times f_s}{2\pi\times \eta}
\end{equation}

\subsection{Eliminating the L-STF}

\begin{figure}[t!]
    \centering
    \includegraphics[width=0.5\textwidth]{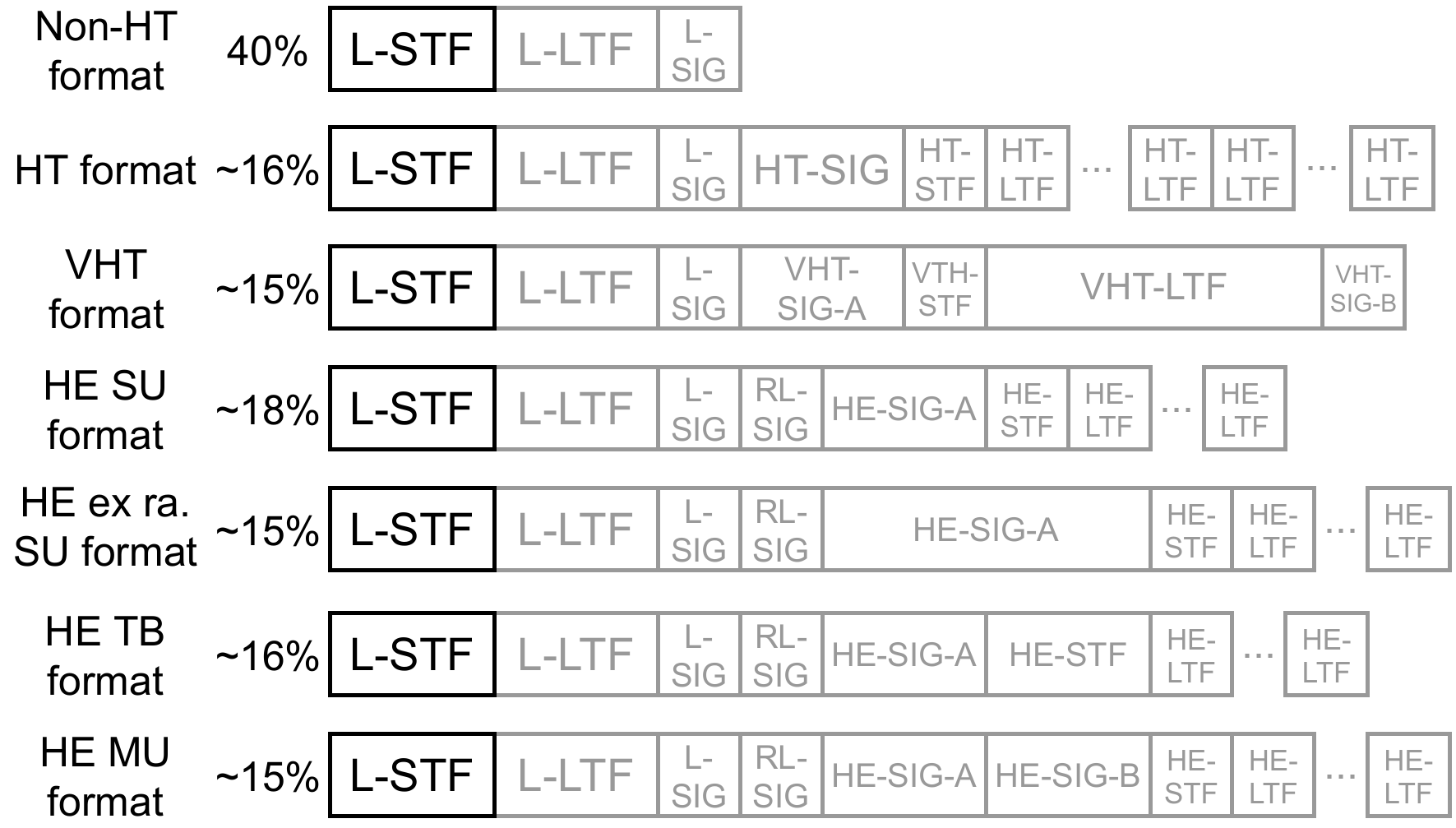}
    \caption{Different IEEE 802.11 preamble formats and the percentages that the L-STF occupies in each preamble~\cite{standard}.\vspace{-0.3cm}}
    \label{fig:formats}
\end{figure}

The L-STF occupies upto 40\% of the preamble and is considered as an integral component of existing and emerging standards of IEEE 802.11 WiFi, as represented in Fig.~\ref{fig:formats}.

Apart from the L-STF, the legacy long training field (L-LTF) plays a key role in fine time synchronization, fine CFO estimation, and channel estimation. Hence, it is also included in all the preamble formats seen in Fig.\ref{fig:formats}. However, the L-LTF has a different pattern with longer training symbol period, and cannot substitute for the L-STF using classical signal processing blocks. In this regard, we propose to use custom-designed neural networks to exploit the intrinsic patterns within the {\em L-LTF symbols} for packet detection and coarse CFO estimation. This obviates the need for L-STF and suggests that it could be completely removed from the preamble.
\section{PRONTO System Description}
\label{sec:proposed}

In this section, we present the details of our proposed PRONTO framework that performs packet detection and coarse CFO estimation without the L-STF. First, we discuss L-LTF extraction which is used to compose the training, validation, and test sets for PRONTO. Then, we explain the CNN-based PRONTO modules for packet detection and coarse CFO estimation, respectively. Finally, we propose a generalized decision flow to make PRONTO compatible with coarse synchronization of standard waveforms.


\subsection{L-LTF Extraction}\label{sec:lltf-extraction}
Consider a PRONTO-enabled waveform that does not contain the L-STF. In this case, PRONTO buffers the received signal in I/Q format and streams these samples to the packet detection module. The objective of this module is to detect potential L-LTF signals that may be embedded somewhere within the streaming samples. PRONTO uses CNNs for both packet detection and coarse CFO estimation, therefore we first describe the procedure for training these models. To setup a training/test pipeline using an OTA standard WiFi dataset, we need to create the labeled training/test datasets by extracting coarsely time-synchronized (detected) L-LTF signals denoted as $X(t)$s. To do this, the start of each packet in the standard dataset is detected through the traditional method (Section~\ref{sec:traditional-packet-detect}), and depending on the waveform in use we extract L-LTF signal, $X(t)$, with known length and known start index, $l$, in the standard waveform. 
For the input data provided to the PRONTO coarse CFO estimation module, we label each $X(t)$ with the coarse CFO, $f$, for the associated packet estimated through the traditional method using the corresponding L-STF (Section~\ref{sec:traditional-packet-detect}). The dataset of $X(t)$ signals recorded and labeled in this manner is later partitioned into training, validation, and test sets.


\subsection{PRONTO Packet Detection Module}
\label{sec:proposed-packet}

\begin{figure*}
    \centering
    \includegraphics[width=0.98\linewidth]{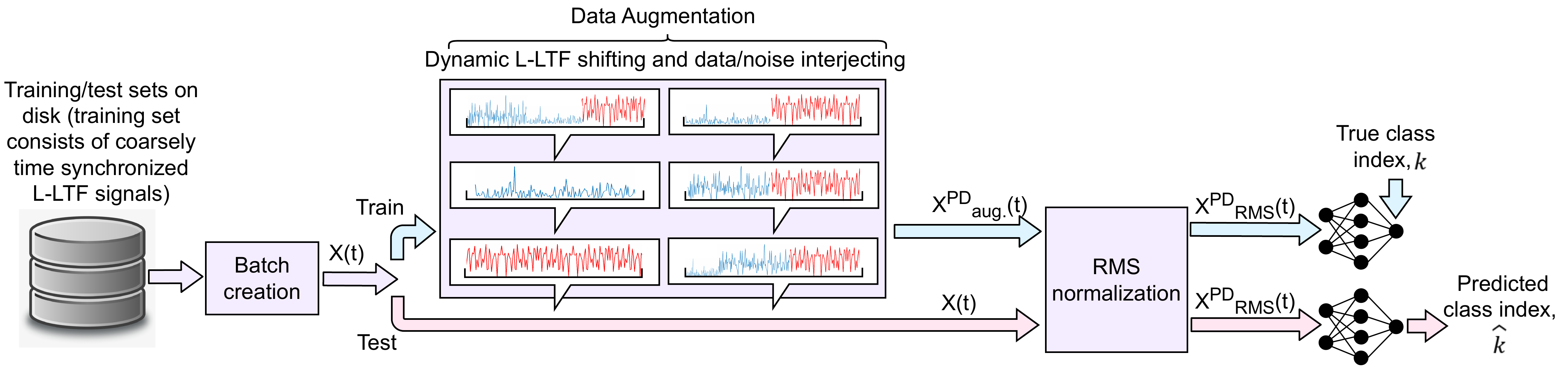}
    \caption{PRONTO training/test pipeline for packet detection as a classification problem, with dynamic L-LTF shifting block in the training path.}\vspace{-0.5cm}
    \label{fig:proposed-packet}
\end{figure*}

\begin{figure}
    \centering
    \includegraphics[width=0.4\textwidth]{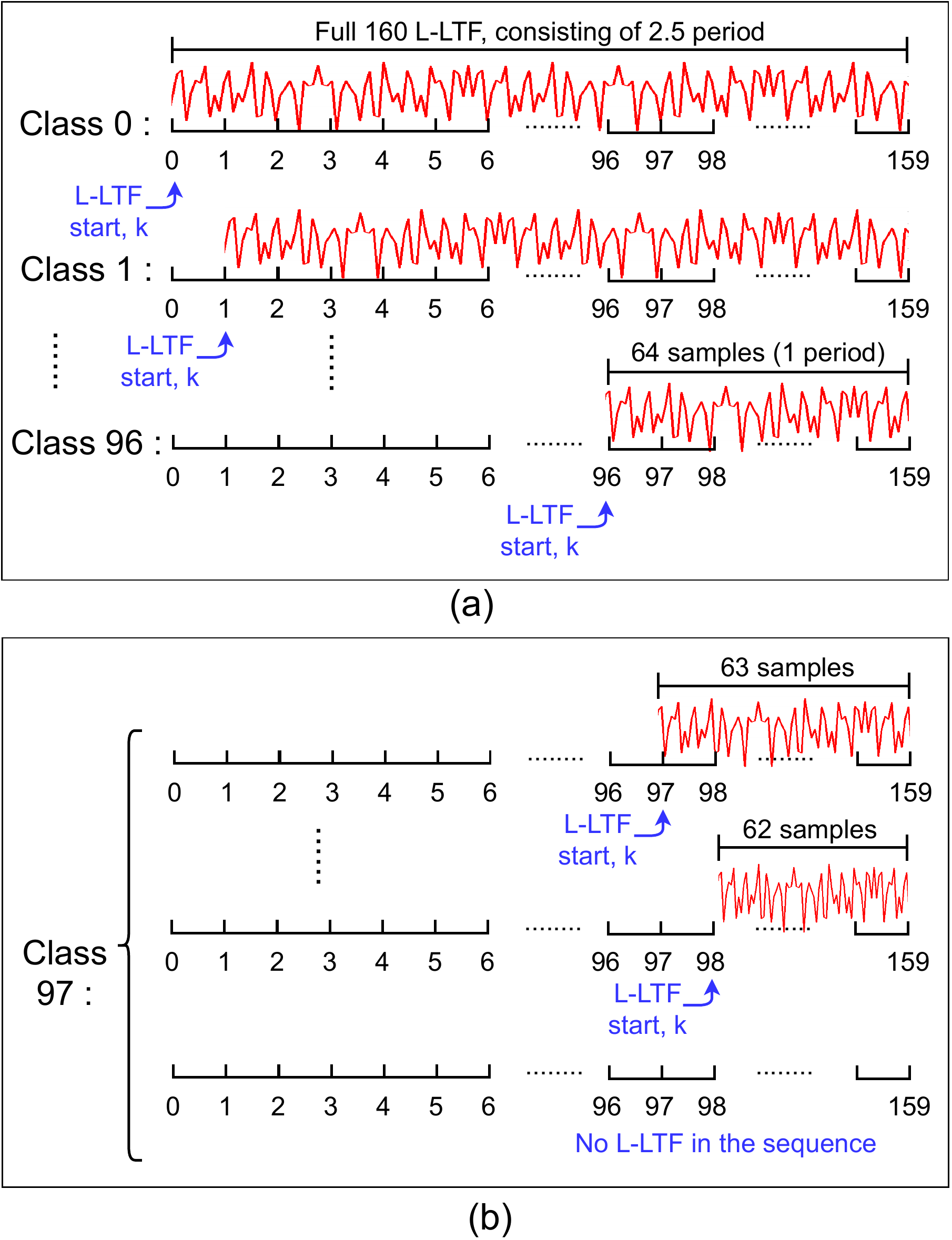}
    \caption{The position of L-LTF in the 160 sample NN input for different packet detection classes. (a) Classes 0 to 96 correspond to the index of packet start, where we have 1 period of L-LTF signal in the sequence. (b) Class 97 shows different cases where less than a period of L-LTF exists in the sequence and the L-LTF pattern cannot be detected.\vspace{-0.5cm}}
    \label{fig:proposed-classes}
\end{figure}

\begin{figure*}
    \centering
    \includegraphics[width=0.82\linewidth]{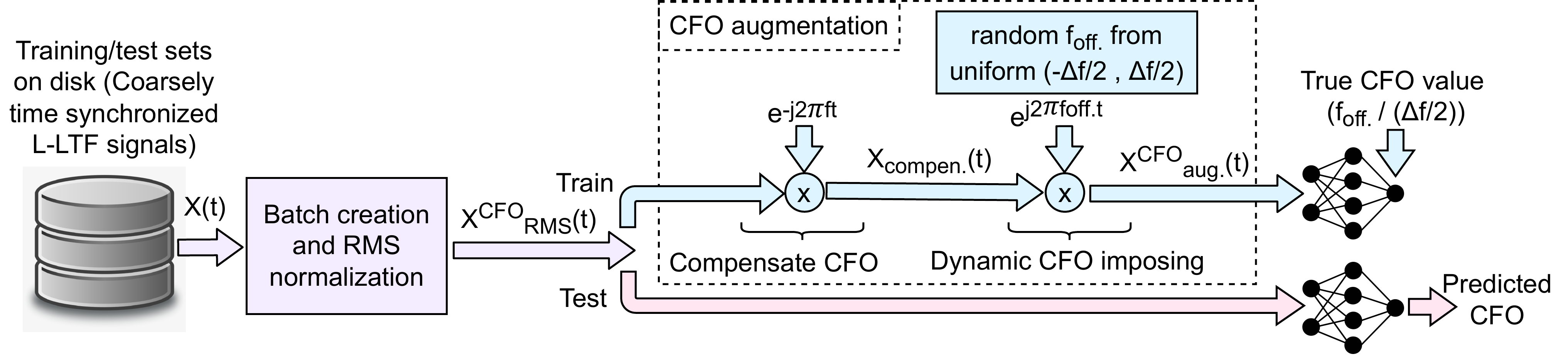}
    \caption{PRONTO training/test pipeline for CFO estimation as a regression problem, with CFO augmentation block in the training path.\vspace{-0.3cm}}
    \label{fig:proposed-cfo}
\end{figure*}

In the first module of PRONTO scheme, we propose a CNN-based solution to detect packets by detecting their L-LTF signals. Our solution can be generalized to detect any other types of signals (e.g., proprietary L-LTF sequences with customized patterns and periods) as long as a specific pattern exists within them. Since the role of packet detection module is to find the start index of a specific pattern in a sequence of time domain samples with \emph{discrete} indices, we model packet detection as a \emph{multi-class classification} problem, which we solve using a deep CNN. The input and output dimensions of the CNN, and the details of data flow in the packet detection module are shown in Fig.~\ref{fig:proposed-packet} and explained in the following.

\noindent \textbf{CNN Input Dimensions.} Inputs of the packet detector CNN are sequences of length $\mathcal{L}$ that contain shifted versions of the L-LTF signal, $X(t)$, in the time domain. To determine the size of the input sequence, we start by studying the L-LTF length. In an OFDM system, the L-LTF signal consists of 2.5 long training symbols, each with length equal to the FFT length, $\mathcal{N}$. The smallest input sequence with length $\mathcal{L}$ to fit one full L-LTF signal (i.e., start index = 0 with respect to the input sequence) is calculated as $\mathcal{L} = 2.5\times \mathcal{N}$. 
We prepare the inputs for the CNN by separating their real and imaginary components and forming inputs of dimensions ($\mathcal{L},2$).

\noindent \textbf{CNN Output Dimensions.} To model packet detection as a classification problem, we map \emph{detection} or \emph{no detection} of a packet to certain classes. 
Each input sequence with length $\mathcal{L}$ is classified as detection classes, if a portion of L-LTF signal, $X(t)$, exists within it. In cases where a portion of $X(t)$ is present, we associate different classes with different start indices of $X(t)$ with respect to the input with length $\mathcal{L}$.
We denote the start index (shift) of $X(t)$ in the input sequence of length $\mathcal{L}$, as $k$, which ranges from 0 to a specific number $K$ (i.e., $k \in \{0 ,..., K\}$). For the CNN to be able to distinguish an L-LTF pattern from a generic pattern of bits, there should be at least one period of the L-LTF in the input. Therefore, $K$ is defined as the largest L-LTF shift in the range [0, $\mathcal{L}$-1], wherein at least $\mathcal{N}$ samples of L-LTF are in the input sequence of length $\mathcal{L}$. $K$ is calculated as $K = \mathcal{L}- \mathcal{N}$.

We associate each L-LTF shift $k$ ($k \in \{0 ,..., K\}$) with one class, and hence, the set of classes is given as $\mathbb{I}^{\PD} = \{0,\cdots, K\}$, where $\mathrm{len}(\mathbb{I}^{\PD}) = (K+1)$.
Apart from these $K+1$ classes where the L-LTF pattern can be detected in the input, there is one additional class that includes all the cases where the L-LTF pattern cannot be recognized. This class includes cases where L-LTF exists in the input but it cannot be detected because $k>K$, as well as the cases where the input does not contain any part of the L-LTF. Hence, we revise the set of classes as  $\mathbb{I}^{\PD} = \mathbb{I}^{\PD} \bigcup \{(K+1)\}$, and the total number of classes are updated to  $\mathrm{len}(\mathbb{I}^{\PD}) = (K+2)$.
Finally, the basis vector (or one hot representation) of the class set $\mathbb{I}^{\PD}$ is represented as $Y^{\PD}\in \{0,1\}^{\mathrm{len}(\mathbb{I}^{\PD})}$.

Fig.~\ref{fig:proposed-classes} shows an example of defining the classes with L-LTF of length $\mathcal{L}=160$, that occurs in 5 MHz, 10 MHz, and 20 MHz bandwidths. In Fig.~\ref{fig:proposed-classes}(a), the start of L-LTF, $k$, in the input sequence is shifted between 0 and $K$ to create $K+1$ classes, where the L-LTF pattern can be detected in the input, giving rise to classes indexed as 0 to 96. Fig.~\ref{fig:proposed-classes}(b) shows different cases where the L-LTF pattern cannot be detected in the input, and hence, the packet is not detected. All these cases are labeled as class 97. Therefore, total number of classes for $\mathcal{L}=160$ is calculated as 98.

\noindent \textbf{Data Augmentation.}
We load to the memory the L-LTF signals, $X(t)$s, of the training set, batch-by-batch. Next, we pass each training $X(t)$ through a \emph{data augmentation} function for packet detection, denoted as $F^{\PD}_{\text{aug.}}(.)$, to dynamically generate different inputs and their corresponding true classes for the packet detector CNN. The goal of the data augmentation block is to artificially create different shift variations in the training input to the CNN, so that the trained packet detector CNN can detect L-LTF signals in different settings with a variety of time shifts and added noise. In this regard, the augmentation function performs a series of steps including shifting the input L-LTFs, as well as inserting noise and random data portions along the shifted L-LTF, to increase resiliency of the classifier, especially for unseen channels and conditions.

First, the mean of $X(t)$ is calculated as $\mu_{X(t)}$.
Second, an empty sequence of $X^{\PD}_{\text{aug.}}(t)$ is created with the same dimensions as those of $X(t)$. 
Third, the L-LTF signal $X(t)$ is shifted $k$ indices ($k \in \{0,...,K\}$) and inserted in $X^{\PD}_{\text{aug.}}(t)$ (that has length $\mathcal{L}$), as shown in Fig.~\ref{fig:proposed-classes}. Fourth, the power of $X(t)$ is calculated as $P_{X(t)}=\frac{1}{\mathcal{L}}|X(t)|^2$, and a random vector with random length in range [0,$k$], is drawn from complex Gaussian distribution with mean=$\mu_{X(t)}$ and variance=$P_{X(t)}$. This vector emulates the potential random data with the same power as L-LTF, and is inserted in $X^{\PD}_{\text{aug.}}(t)$ alongside the shifted $X(t)$. 
Fifth, the power of noise $P_{N(t)}$ in the L-LTF signal is calculated, and if there are empty periods left in $X^{\PD}_{\text{aug.}}(t)$, they are filled with random values drawn from complex Gaussian distribution with mean $\mu_{X(t)}$ and variance $P_{N(t)}$.
The function $F^{\PD}_{\text{aug.}}(.)$ returns $X^{\PD}_{\text{aug.}}(t)$ as well as the shift of L-LTF, $k$, as the true class index, as shown in (\ref{eq:aug_pd}). The data augmentation block is highlighted in Fig.~\ref{fig:proposed-packet}.
\vspace{-0.3cm}

\begin{equation}\label{eq:aug_pd}
    X^{\PD}_{\text{aug.}}(t)~,~k = F^{\PD}_{\text{aug.}}(X(t),\ P_{N(t)},\ P_{X(t)})
\end{equation}

\noindent \textbf{RMS Normalization.} At this stage our augmented signal $X^{\PD}_{\text{aug.}}(t)$ contains the shifted $X(t)$ along with periods of data and noise with different amplitudes.
In order for the packet detector CNN to be able to detect signals in different environments with different amplitudes, we perform a root mean square (RMS) normalization for each augmented signal $X^{\PD}_{\text{aug.}}(t)$, as in (\ref{eq:rms}).
\vspace{-0.2cm}

\begin{equation}\label{eq:rms}
    X^{\PD}_{\text{RMS}}(t) = \frac{X^{\PD}_{\text{aug.}}(t)}{\sqrt{\frac{1}{\mathcal{L}}\sum|X^{\PD}_{\text{aug.}}(t)|^2}}
\end{equation}

The RMS normalization block that follows the data augmentation block is shown in Fig.~\ref{fig:proposed-packet}.

\noindent \textbf{CNN Modeling.}
Recall that real and imaginary components of $X^{\PD}_{\text{RMS}}(t)$ are separated to form an input of dimensions ($\mathcal{L},2$), which is used for training and testing the CNN along with one hot representation, $Y^{\PD}\in \{0,1\}^{\mathrm{len}(\mathbb{I}^{\PD})}$, of the true class index. The proposed CNN-based classifier predicts the occurrence probability of each class, as in (\ref{eq:s_PD}), where $\gamma$ is \emph{Softmax} and $F_{\theta}^{\PD}(.)$ denotes the classifier CNN.
\begin{equation} \label{eq:s_PD}
\hat{Y}^{\PD} =\gamma\,(F_{\theta}^{\PD}(X^{\PD}_{\text{RMS}}(t))) ~~~~~~F_{\theta}^{\PD}:\mathbb{R}^{\mathcal{L}\times 2} \mapsto \mathbb{R}^{\mathrm{len}(\mathbb{I}^{\PD})}
\end{equation}

As shown in Fig.~\ref{fig:proposed-packet}, we train the CNN with $X^{\PD}_{\text{RMS}}(t)$ using \emph{catergorical cross-entropy} loss for several hundreds of epochs, to create millions of variations in the L-LTF pattern, as the shifts and data and/or noise chunks happen randomly for each $X(t)$ in each epoch. 
During the test phase, we do not need the augmentation block described in (\ref{eq:aug_pd}), as the sampled wireless input by nature resembles the augmented L-LTF sequences in Fig.~\ref{fig:proposed-packet}. Therefore, we only process the test sequences through the RMS normalization block in (\ref{eq:rms}) and feed them to the trained CNN model.

The predicted class index, $\hat{k}$, is determined using the probability vector, $\hat{Y}^{\PD}$, as in (\ref{eq:argmax}).
\begin{equation}\label{eq:argmax}
    \hat{k} = \underset{\hat{k} \in \{0,...,\,\mathrm{len}(\mathbb{I}^{\PD})-1\}}{\arg\max}~(\hat{Y}^{\PD})
\end{equation}


\subsection{PRONTO Coarse CFO Estimation Module} \label{sec:proposed-cfo}
The second module of PRONTO is designed to estimate coarse CFO using the L-LTF signals, once a packet is detected in the system. We model coarse CFO estimation as a \emph{regression} problem where the intrinsic properties within the L-LTF patterns are exploited by a CNN to estimate the coarse CFO (see Fig.~\ref{fig:proposed-cfo}).

\noindent \textbf{CNN Input and Output Dimensions.} The inputs to the coarse CFO estimator CNN are batches of L-LTF signals, $X(t)$s, of dimensions ($\mathcal{L},2$) where I and Q are passed through 2 separate channels to form the second dimension. Since we consider coarse CFO estimation as a regression problem, the label space of the PRONTO CFO estimator CNN converges to a single valued prediction represented by 1 neuron ($\mathrm{len}(\mathbb{I}^{\CFO})=1$).




\noindent \textbf{RMS Normalization.}
To start with training, validation, and test phases, we normalize each L-LTF signal, through RMS normalization as in (\ref{eq:cfo-rms}).
\vspace{-0.3cm}

\begin{equation}\label{eq:cfo-rms}
    X^{\CFO}_{\text{RMS}}(t) = \frac{X(t)}{\sqrt{\frac{1}{\mathcal{L}}\sum|X(t)|^2}}
\end{equation}

The RMS normalization step in the training/test pipeline is shown in Fig.~\ref{fig:proposed-cfo}.

\noindent \textbf{Data Augmentation (CFO Augmentation).}
For training a robust CFO estimator, different variations of the input perturbed with different CFO values should be provided to the CNN. To do this, we insert an intermediate CFO augmentation block in the training pipeline, as shown in Fig.~\ref{fig:proposed-cfo}.

Before adding \emph{augmenting} CFO values to the normalized signal, $X^{\CFO}_{\text{RMS}}(t)$, the inherent coarse CFO, $f$, of the signal must be compensated. This $f$ is previously recorded for each L-LTF during L-LTF extraction, and is required only during the training process, as shown in Fig.~\ref{fig:proposed-cfo}. The inherent coarse CFO, $f$, is compensated on $X^{\CFO}_{\text{RMS}}(t)$ with sampling frequency $f_s$, as given in (\ref{eq:compensate}). Here, $l$ is the L-LTF start index in the original waveform as defined in Section~\ref{sec:lltf-extraction}. 
\vspace{-0.5cm}

\begin{equation}\label{eq:compensate}
    X_{\text{compen.}}(t) = X^{\CFO}_{\text{RMS}}(t)\times e^{-j2\pi ft/f_s} , t=l,\ldots,l+\mathcal{L}-1
\end{equation}

After compensating the CFO, $f$, an augmenting CFO needs to be applied to $X_{\text{compen.}}(t)$.
We choose our augmenting CFO to be in range [$-\frac{\Delta f}{2}, \frac{\Delta f}{2}$], where $\Delta f$ is sub-carrier frequency spacing. In the augmentation step, we draw a random CFO value, $f_{\text{off.}}$, from a uniform distribution given as $f_{\text{off.}}\sim U(-\frac{\Delta f}{2} , \frac{\Delta f}{2})$, and apply it to the signal $X_{\text{compen.}}(t)$, as shown in (\ref{eq:augmented}).
\vspace{-0.5cm}

\begin{equation}\label{eq:augmented}
    X^{\CFO}_{\text{aug.}}(t) = X_{\text{compen.}}(t)\times e^{j2\pi f_{\text{off.}}t/f_s},~~ t=0,\ldots,\mathcal{L}-1
\end{equation}

The augmentation function $F^{\CFO}_{\text{aug.}}(x)$ is defined as a cascade of (\ref{eq:compensate}) and (\ref{eq:augmented}), with the assumption that $l=\mathcal{L}$ (i.e., the L-STF and L-LTF have the same length) as in (\ref{eq:f_aug}).
\begin{equation}\label{eq:f_aug}
    F^{\CFO}_{\text{aug.}}(x) = x\times e^{j2\pi (f_{\text{off.}}(t-\mathcal{L})-ft)/f_s}, t=\mathcal{L},\ldots,2\mathcal{L}-1
\end{equation}

During augmentation, for each $X^{\CFO}_{\text{aug.}}(t)$, $f_{\text{off.}}$ is recorded as the true CFO label. For a regression problem, the labels need to be normalized, as well. In our case, $f_{\text{off.}}$ is scaled to the range of [-1,1] as in (\ref{eq:y_normalize}) to generate true label $Y^{\CFO}$ for the CNN.
\vspace{-0.3cm}
\begin{equation}\label{eq:y_normalize}
    Y^{\CFO} = \frac{f_{\text{off.}}}{\Delta f/2}
\end{equation}

The function $F^{\CFO}_{\text{aug.}}(.)$ returns the augmented L-LTF signal, $X^{\CFO}_{\text{aug.}}(t)$, along with its corresponding normalized true label, $Y^{\CFO}$, as in (\ref{eq:f-cfo-return}).

\begin{equation}\label{eq:f-cfo-return}
    X^{\CFO}_{\text{aug.}}(t)~,~Y^{\CFO} = F^{\CFO}_{\text{aug.}}(X^{\CFO}_{\text{RMS}}(t))
\end{equation}

\noindent \textbf{CNN Modeling.} 
Recall that real and imaginary components of $X^{\CFO}_{\text{aug.}}(t)$s are separated to form inputs with dimensions ($\mathcal{L},2$) which are fed to the training function along with their corresponding true labels, $Y^{\CFO}$s. We design PRONTO coarse CFO estimator to learn to map each augmented L-LTF signal, $X^{\CFO}_{\text{aug.}}(t)$, to a true CFO value, $Y^{\CFO}$, as in (\ref{eq:y_CFO}) where $\delta$ is \emph{tanh}, $F_{\theta}^{\CFO}(.)$ denotes the regressor CNN, and $\hat{Y}^{\CFO}$ represents the predicted coarse CFO.
\begin{equation} \label{eq:y_CFO}
\hat{Y}^{\CFO} =\delta\,(F_{\theta}^{\CFO}(X^{\CFO}_{\text{aug.}}(t))),~~~~~~F_{\theta}^{\CFO}:\mathbb{R}^{\mathcal{L}\times 2} \mapsto  \mathbb{R}^1
\end{equation}

We train the CFO estimator CNN with batches of $X^{\CFO}_{\text{aug.}}(t)$ signals. The \emph{mean squared error} loss is calculated between $Y^{\CFO}$ and $\hat{Y}^{\CFO}$ for each batch. We train the CNN for several hundreds of epochs with millions of CFO variations introduced in different L-LTF signals.

In the test phase, $X(t)$ signals only need to be normalized using (\ref{eq:rms}). 

After the test phase, the actual predicted coarse CFO, $\hat{f}_{\text{off}}$, is calculated as in (\ref{eq:f_hat}), where $\hat{Y}^{\CFO}$ is the CNN prediction in range [-1,1].
\vspace{-0.3cm}

\begin{equation}\label{eq:f_hat}
    \hat{f}_{\text{off}} = \hat{Y}^{\CFO} \times \Delta f/2
\end{equation}


\subsection{System Overview} \label{sec:proposed-system}
\begin{figure}
    \centering
    \includegraphics[width=0.35\textwidth]{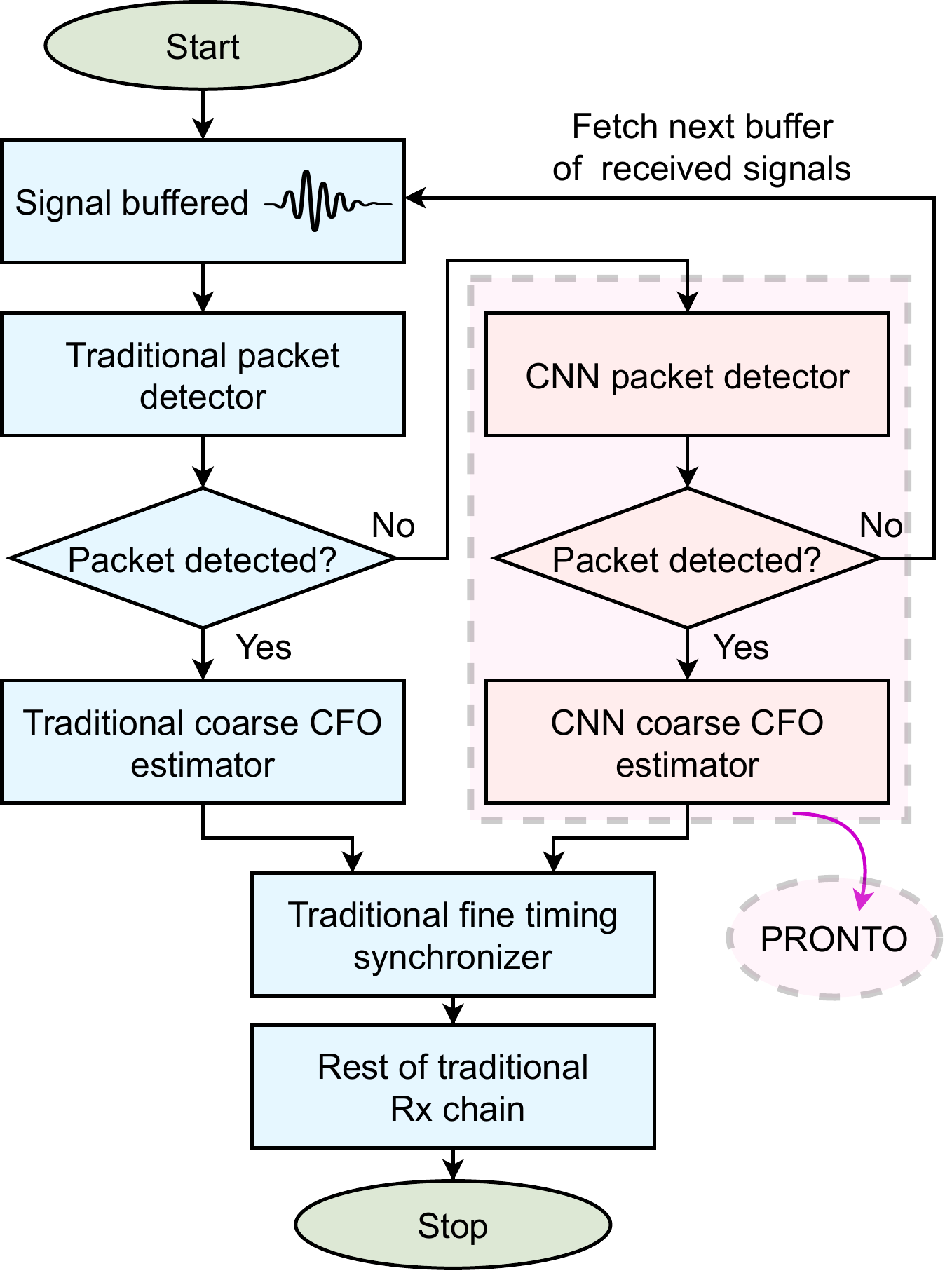}
    \caption{Generalized decision flow for making PRONTO compatible with standard IEEE 802.11 waveforms. The system first looks for the L-STF in the received signal. If no L-STF is detected, it sends the signal to PRONTO packet detector CNN for L-LTF search.\vspace{-0.3cm}}
    \label{fig:flow-chart}
\end{figure}

We propose an adaptive decision flow to make PRONTO compatible with standard IEEE 802.11 waveforms (i.e., containing both L-STF and L-LTF in the preamble), as well as the proposed modified version of this waveform (containing no L-STF in the preamble), as shown in Fig.~\ref{fig:flow-chart}.
First the received signal is buffered and sent to the traditional packet detector block. If the latter detects the packet, it means that the L-STF pattern exists in the received signal, and the signal goes through the traditional processing chain. If the traditional packet detector cannot detect a packet, there is a chance that the buffered signal is a modified waveform that starts with L-LTF. Therefore, we send the buffered signal to PRONTO packet detector to search for L-LTF pattern. If the L-LTF pattern is detected, we proceed with the PRONTO coarse CFO estimator. After coarse CFO estimation, the packet goes through the rest of the traditional receiver chain. If PRONTO packet detector CNN cannot detect the L-LTF pattern, it goes up to fetch the next buffered set of I/Q samples of the received signal. The whole process iterates again starting with the traditional packet detector. 
\section{Datasets and Deep Learning Setup}
\label{sec:dataset}

\subsection{Dataset Description}
We use two OTA WiFi datasets to evaluate PRONTO.

\subsubsection{Oracle Dataset} We use a publicly available\footnote{https://genesys-lab.org/oracle} OTA WiFi dataset~\cite{oracle-kunal} that is referred to as \emph{Oracle} in the rest of this paper. As described in the metadata files of this dataset, signals are collected in an indoor environment from 16 different USRP X310 software defined radios (SDRs) that transmit IEEE 802.11 Non-HT frames (see Fig.~\ref{fig:formats}) with OFDM modulation, QPSK 3/4 in 2.4 GHz frequency with 5 MHz bandwidth. The distance between the transmitter and receiver varies between 2 ft and 62 ft with steps of 6 ft (11 different distances). The wireless channel is continuously sampled using a receiver USRP B210. We set the parameters defined in Section~\ref{sec:proposed}, using the dataset properties, as shown in Table~\ref{tab:parameters}.

\subsubsection{Arena Dataset} To show how pre-trained PRONTO CNNs perform in new environments, we use the dataset used in~\cite{nasim-spinn,bahar-spinn}, which we refer to as \emph{Arena} in the rest of this paper. As the name suggests, this dataset is collected in Arena~\cite{arena}, an indoor lab environment with X310 SDR arrays on the ceiling. A pair of X310 SDRs are used to collect $\sim$152k IEEE 802.11a packets with 16QAM modulation at 10 MHz bandwidth. The L-LTF parameters of this dataset are the same as Oracle dataset described in Table~\ref{tab:parameters}, except for the sampling frequency ($f_s$=10 MHz) and the sub-carrier frequency spacing ($\Delta f$=156.5 kHz). As the SDR locations are fixed, transmission power is intentionally varied during data collection to emulate different SNRs.

\subsection{L-LTF Extraction} \label{sec:setup-extraction}
For extracting L-LTF signals from each dataset, we find packet offsets using the traditional packet detection method explained in Section~\ref{sec:traditional-packet-detect} and implemented in \texttt{MATLAB} function \texttt{wlanPacketDetect} with $\beta$=0.8.
The packet offset in the standard IEEE 802.11 waveform is the offset from the beginning of the L-STF. We add an offset of 160 (L-STF length) to it and extract the next 160-sample sequence as coarsely time-synchronized L-LTF.
To calculate the noise power, $P_{N(t)}$, in each L-LTF signal that is needed for the packet detection problem as shown in (\ref{eq:aug_pd}), we use \texttt{MATLAB} function \texttt{helperNoiseEstimate}. To obtain inherent coarse CFO, $f$, used in (\ref{eq:compensate}), we record coarse CFOs estimated using the L-STF with the traditional coarse CFO estimation method described in Section~\ref{sec:traditional-cfo}. This step is implemented in \texttt{MATLAB} function \texttt{wlanCoarseCFOEstimate}. We extract $\sim$663k and $\sim$152k L-LTF signals from the recorded sequences in Oracle and Arena datasets, respectively, along with their corresponding noise power, $P_{N(t)}$, and their coarse CFOs, $f$s. These signals form our training, validation, and test sets that are used in packet detection and coarse CFO estimation problems. 

\begin{table}[t!!!]
    \centering
    \resizebox{0.45\textwidth}{!}{
    \begin{tabular}{lcr}
         \hline\hline\xrowht[()]{5pt}
         Parameter & Notation & Value \\
         \hline
         \xrowht[()]{5pt}
         Length of FFT (Length of& \multirow{2}{*}{$\mathcal{N}$} & \multirow{2}{*}{64} \\
          one L-LTF training symbol)  &&\\
         \hline
         \xrowht[()]{5pt}
         Full length of L-LTF & $\mathcal{L}$ & 160\\
         \hline
         \xrowht[()]{5pt}
         Start index of L-LTF in the & \multirow{2}{*}{$l$} & \multirow{2}{*}{160}\\
          original waveform &&\\
         \hline
         \xrowht[()]{5pt}
         Output layer size for classification & $\mathrm{len}(\mathbb{I}^{\text{PD}})$ & 98\\
         \hline
         \xrowht[()]{5pt}
         Output layer size for regression & $\mathrm{len}(\mathbb{I}^{\text{CFO}})$ & 1\\
         \hline
         \xrowht[()]{5pt}
         Sampling frequency & $f_s$ & 5 MHz\\
         \hline
         \xrowht[()]{5pt}
         Sub-carrier frequency spacing & $\Delta f$ & 78.125 kHz\\
         \hline\hline
    \end{tabular}
    }
    \caption{Parameters for 5 MHz WiFi signals \cite{ieee-book} in the Oracle dataset\cite{oracle-kunal}.}
    \label{tab:parameters}
\end{table}

\subsection{Deep Learning Setup}

\begin{figure}
    \centering
    \includegraphics[width=0.5\textwidth]{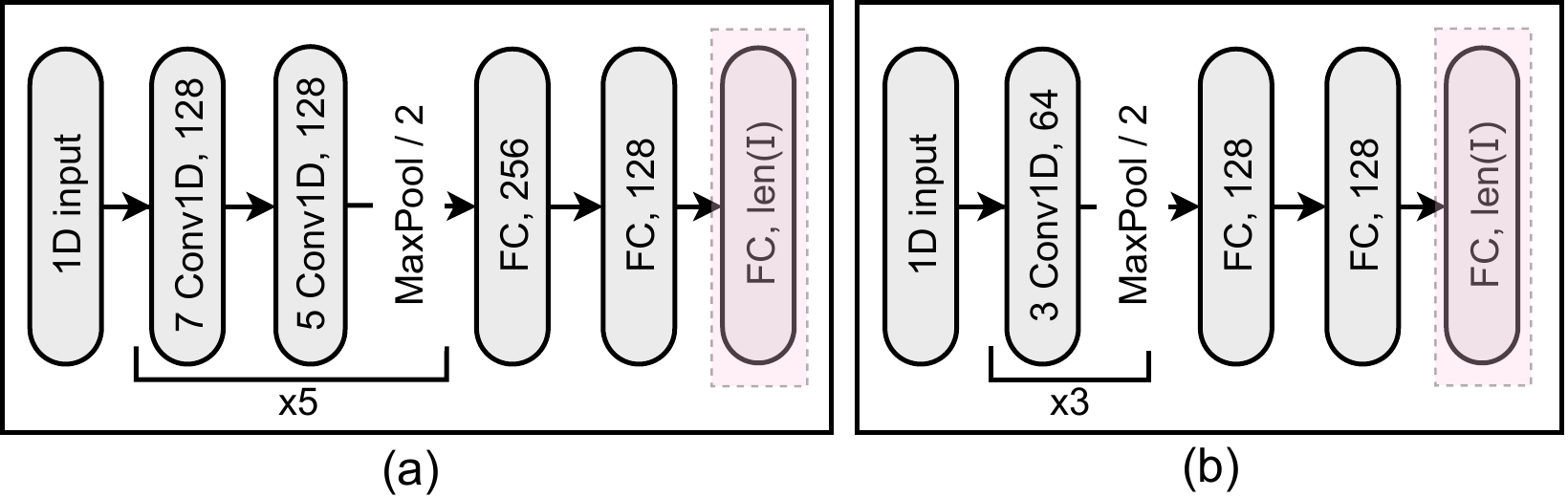}
    \caption{(a) PRONTO-L with $\sim$1M parameters, and (b) PRONTO-S with $\sim$200k parameters, used for packet detection and coarse CFO estimation. The highlighted last layer has size 98 with \emph{Softmax} activation for packet detection (classification), and size 1 with \emph{tanh} activation for coarse CFO estimation (regression).\vspace{-0.3cm}}
    \label{fig:nn-arch}
\end{figure}

\noindent \textbf{Data Augmentation.}
For the deep learning framework\footnote{https://github.com/nasimsoltani/PRONTO}, we use Keras~\cite{keras}. We use a data generator class inherited from \texttt{keras.utils.Sequence}, a special class from Keras libraries that gives us the possibility of loading the data to the memory batch-by-batch. This helps in efficiently performing the described data augmentation steps during packet detection and coarse CFO estimation. 

\noindent \textbf{CNN Input Dimensions.} For both problems, we configure the input size of the CNN as $\mathcal{L}$ = 160, which is the size of an L-LTF signal. We separate real and imaginary parts in the L-LTF signal to form inputs of size (160, 2) with I/Q separated in the last dimension as separate channels.

\noindent \textbf{CNN Architectures.} We compare the performance of two 1D forward CNN architectures, (i) a large CNN called PRONTO-L with 13 layers and $\sim$1M parameters, and (ii) a smaller CNN called PRONTO-S with 6 layers and $\sim$200k parameters. Both architectures are combinations of 1D convolutional layers, 1D MaxPooling layers, and fully connected (FC) layers. The details of PRONTO-L and PRONTO-S architectures are shown in Fig.~\ref{fig:nn-arch}. In this figure, the convolution layers are characterized with two numbers. The first number is the convolution kernel (filter) size, and the second number indicates the number of kernels in the layer. For example, \emph{5 Conv1D, 128} means a convolution layer with 128 kernels of size 7 each. The FC layers are characterized with one value that is the output size of the FC layer. The MaxPooling layers have window size of 2 with stride 2.

\noindent \textbf{CNN Output Dimensions.}
The final layer of CNNs in Fig.~\ref{fig:nn-arch} is an FC layer whose size and activation function changes depending on the studied problem. For packet detection that is considered a classification problem with 98 classes, the final layer is of size 98 with \emph{Softmax} activation. For CFO estimation that is considered a regression problem, the final layer has only one neuron with \emph{tanh} activation function.

The PRONTO-L and PRONTO-S architectures shown in Fig.~\ref{fig:nn-arch}, with the aforementioned input and output dimensions have $\sim$1 million and $\sim$200 thousand parameters, respectively.
\section{Evaluations}\label{sec:eval}
In this section, we validate PRONTO in terms of both packet detection and coarse CFO estimation performance using the OTA datasets described in Section~\ref{sec:dataset}. We use Oracle dataset to evaluate PRONTO training/test performance as the primary step. We further show how PRONTO trained in this initial environment (Oracle dataset) performs in a new environment (Arena dataset). Moreover, we present computational complexity and run-time of PRONTO on 4 different platforms and compare our results with the closest related work to PRONTO~\cite{packet-detection,packet-detect-cfo,coarse-cfo}. 

\begin{figure}[t!!!]
    \centering
    \begin{subfigure}[t]{0.23\textwidth}
        \centering
        \includegraphics[width=\linewidth]{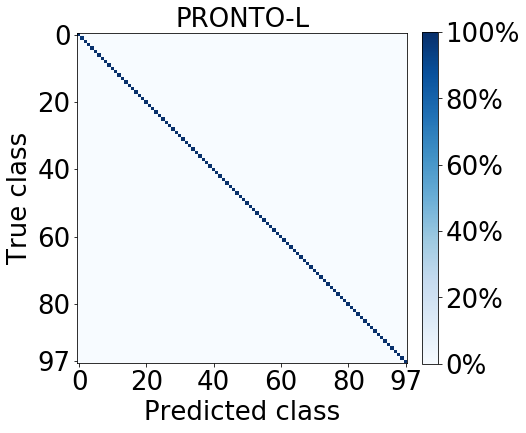}
    \end{subfigure}
    \begin{subfigure}[t]{0.23\textwidth}
    \centering
        \includegraphics[width=\linewidth]{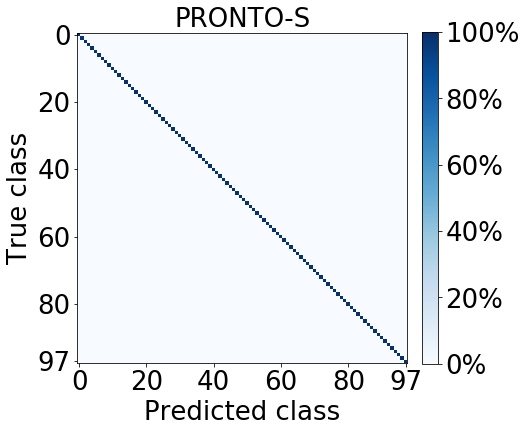}
    \end{subfigure}
    \caption{Confusion matrices for packet detection using L-LTF, with PRONTO-L and PRONTO-S architectures. Both architectures yield 100\% accuracy, therefore, for packet detection we choose PRONTO-S that has 1/\nth{5} parameters compared to PRONTO-L.\vspace{-0.3cm}}
    \label{fig:result-conf-detection}
\end{figure}

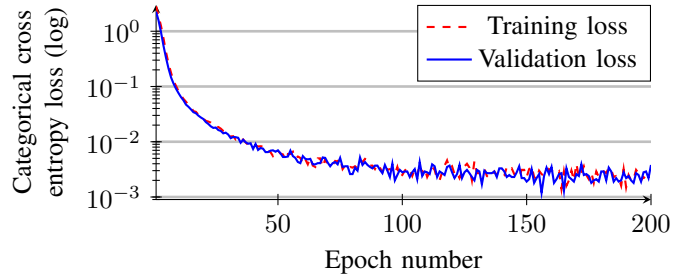
\begin{figure}[t!!!]
\centering
\begin{tikzpicture}
        \begin{axis}[
            axis lines = left,
            xlabel = Epoch number,
            ylabel = {Categorical cross entropy loss (log)},
            y label style = {align=center,text width=3cm},
            ymin=0.0009,
            ymax=3,
            ymode = log,
            xmin=1,
            xmax=200,
            height=0.23\textwidth,
            width=0.45\textwidth,
            legend style={at={(1,1)},
                anchor=north east,legend columns=1},
            ymajorgrids=true,
            grid style={line width=1pt,draw=gray!50},
        ]
        \addplot [thick,color=red, dashed]
        coordinates{ 
        (1,2.8421)(2,1.8944)(3,1.2277)(4,0.7331)(5,0.4172)(6,0.2598)(7,0.1778)(8,0.1260)(9,0.1027)(10,0.0826)(11,0.0695)(12,0.0598)(13,0.0531)(14,0.0468)(15,0.0427)(16,0.0358)(17,0.0348)(18,0.0309)(19,0.0283)(20,0.0262)(21,0.0239)(22,0.0232)(23,0.0222)(24,0.0205)(25,0.0185)(26,0.0185)(27,0.0173)(28,0.0157)(29,0.0149)(30,0.0139)(31,0.0124)(32,0.0126)(33,0.0110)(34,0.0118)(35,0.0105)(36,0.0110)(37,0.0097)(38,0.0088)(39,0.0087)(40,0.0091)(41,0.0086)(42,0.0082)(43,0.0083)(44,0.0082)(45,0.0081)(46,0.0068)(47,0.0065)(48,0.0055)(49,0.0062)(50,0.0069)(51,0.0061)(52,0.0059)(53,0.0048)(54,0.0060)(55,0.0060)(56,0.0066)(57,0.0050)(58,0.0054)(59,0.0051)(60,0.0040)(61,0.0048)(62,0.0039)(63,0.0051)(64,0.0037)(65,0.0035)(66,0.0050)(67,0.0050)(68,0.0035)(69,0.0048)(70,0.0039)(71,0.0040)(72,0.0038)(73,0.0045)(74,0.0044)(75,0.0050)(76,0.0032)(77,0.0037)(78,0.0033)(79,0.0035)(80,0.0034)(81,0.0036)(82,0.0033)(83,0.0044)(84,0.0043)(85,0.0049)(86,0.0046)(87,0.0037)(88,0.0037)(89,0.0032)(90,0.0039)(91,0.0034)(92,0.0028)(93,0.0037)(94,0.0031)(95,0.0032)(96,0.0031)(97,0.0026)(98,0.0035)(99,0.0023)(100,0.0025)(101,0.0030)(102,0.0034)(103,0.0031)(104,0.0024)(105,0.0032)(106,0.0025)(107,0.0038)(108,0.0027)(109,0.0033)(110,0.0023)(111,0.0029)(112,0.0035)(113,0.0031)(114,0.0021)(115,0.0020)(116,0.0028)(117,0.0037)(118,0.0046)(119,0.0033)(120,0.0036)(121,0.0026)(122,0.0028)(123,0.0022)(124,0.0029)(125,0.0036)(126,0.0040)(127,0.0034)(128,0.0020)(129,0.0045)(130,0.0027)(131,0.0027)(132,0.0030)(133,0.0033)(134,0.0024)(135,0.0022)(136,0.0025)(137,0.0029)(138,0.0025)(139,0.0024)(140,0.0022)(141,0.0024)(142,0.0035)(143,0.0031)(144,0.0033)(145,0.0026)(146,0.0024)(147,0.0023)(148,0.0031)(149,0.0031)(150,0.0023)(151,0.0032)(152,0.0036)(153,0.0023)(154,0.0021)(155,0.0030)(156,0.0026)(157,0.0024)(158,0.0024)(159,0.0032)(160,0.0030)(161,0.0028)(162,0.0026)(163,0.0029)(164,0.0014)(165,0.0029)(166,0.0016)(167,0.0020)(168,0.0030)(169,0.0020)(170,0.0040)(171,0.0023)(172,0.0028)(173,0.0036)(174,0.0025)(175,0.0021)(176,0.0028)(177,0.0018)(178,0.0029)(179,0.0023)(180,0.0026)(181,0.0029)(182,0.0027)(183,0.0034)(184,0.0031)(185,0.0023)(186,0.0024)(187,0.0022)(188,0.0016)(189,0.0032)(190,0.0015)(191,0.0025)(192,0.0020)(193,0.0021)(194,0.0031)(195,0.0027)(196,0.0020)(197,0.0023)(198,0.0021)(199,0.0020)(200,0.0018)
        };
        \addlegendentry{Training loss}
        \addplot [thick,color=blue]
        coordinates{ 
        (1,2.2557)(2,1.5576)(3,0.9465)(4,0.5388)(5,0.3283)(6,0.2114)(7,0.1466)(8,0.1132)(9,0.0913)(10,0.0761)(11,0.0644)(12,0.0562)(13,0.0471)(14,0.0430)(15,0.0386)(16,0.0343)(17,0.0328)(18,0.0297)(19,0.0275)(20,0.0258)(21,0.0227)(22,0.0210)(23,0.0190)(24,0.0178)(25,0.0181)(26,0.0163)(27,0.0157)(28,0.0142)(29,0.0146)(30,0.0135)(31,0.0116)(32,0.0122)(33,0.0121)(34,0.0118)(35,0.0096)(36,0.0109)(37,0.0098)(38,0.0092)(39,0.0094)(40,0.0074)(41,0.0091)(42,0.0071)(43,0.0077)(44,0.0075)(45,0.0065)(46,0.0065)(47,0.0069)(48,0.0070)(49,0.0066)(50,0.0070)(51,0.0062)(52,0.0053)(53,0.0053)(54,0.0054)(55,0.0069)(56,0.0055)(57,0.0051)(58,0.0043)(59,0.0041)(60,0.0055)(61,0.0049)(62,0.0048)(63,0.0049)(64,0.0039)(65,0.0039)(66,0.0041)(67,0.0041)(68,0.0052)(69,0.0051)(70,0.0042)(71,0.0036)(72,0.0041)(73,0.0043)(74,0.0039)(75,0.0051)(76,0.0035)(77,0.0036)(78,0.0036)(79,0.0026)(80,0.0026)(81,0.0034)(82,0.0030)(83,0.0052)(84,0.0038)(85,0.0043)(86,0.0038)(87,0.0035)(88,0.0033)(89,0.0037)(90,0.0037)(91,0.0031)(92,0.0027)(93,0.0038)(94,0.0030)(95,0.0023)(96,0.0040)(97,0.0020)(98,0.0025)(99,0.0031)(100,0.0028)(101,0.0029)(102,0.0029)(103,0.0023)(104,0.0028)(105,0.0032)(106,0.0021)(107,0.0043)(108,0.0024)(109,0.0029)(110,0.0028)(111,0.0026)(112,0.0039)(113,0.0036)(114,0.0023)(115,0.0029)(116,0.0027)(117,0.0025)(118,0.0035)(119,0.0037)(120,0.0028)(121,0.0022)(122,0.0026)(123,0.0038)(124,0.0022)(125,0.0028)(126,0.0031)(127,0.0027)(128,0.0023)(129,0.0020)(130,0.0031)(131,0.0029)(132,0.0017)(133,0.0030)(134,0.0027)(135,0.0027)(136,0.0027)(137,0.0029)(138,0.0023)(139,0.0033)(140,0.0033)(141,0.0029)(142,0.0023)(143,0.0025)(144,0.0023)(145,0.0023)(146,0.0028)(147,0.0036)(148,0.0022)(149,0.0022)(150,0.0019)(151,0.0021)(152,0.0032)(153,0.0018)(154,0.0022)(155,0.0026)(156,0.0012)(157,0.0024)(158,0.0034)(159,0.0026)(160,0.0035)(161,0.0018)(162,0.0021)(163,0.0026)(164,0.0027)(165,0.0013)(166,0.0033)(167,0.0023)(168,0.0021)(169,0.0019)(170,0.0026)(171,0.0024)(172,0.0039)(173,0.0025)(174,0.0026)(175,0.0024)(176,0.0018)(177,0.0020)(178,0.0021)(179,0.0019)(180,0.0029)(181,0.0029)(182,0.0020)(183,0.0034)(184,0.0029)(185,0.0019)(186,0.0022)(187,0.0019)(188,0.0020)(189,0.0028)(190,0.0022)(191,0.0024)(192,0.0020)(193,0.0018)(194,0.0027)(195,0.0025)(196,0.0019)(197,0.0028)(198,0.0029)(199,0.0022)(200,0.0038)
        };
        \addlegendentry{Validation loss}
        \end{axis}
    \end{tikzpicture}
    \caption{Training and validation loss in packet detection using PRONTO-S.\vspace{-0.6cm}}
    \label{fig:packet-detection-loss}
\end{figure}

\subsection{Packet Detection Evaluation}\label{sec:eval-packet}

We shuffle Oracle L-LTF dataset and partition it into 70\%, 10\%, and 20\% portions to form the training, validation, and test sets, respectively. We train the CNNs PRONTO-L and PRONTO-S shown in Fig.~\ref{fig:nn-arch}, on the training set. 
We stop training when validation accuracy does not improve during 100 consecutive epochs. We test the trained model on the test set, and for each signal in the latter, we calculate the predicted class using (\ref{eq:argmax}). The accuracy over the test set is calculated by dividing the number of correctly predicted signals by the total number of signals in the test set. We observe that both PRONTO-L and PRONTO-S, each with 98 classes defined in Fig.~\ref{fig:proposed-classes}, can detect packets with 100\% accuracy using only L-LTF signals. Fig.~\ref{fig:result-conf-detection} shows the confusion matrices of PRONTO-L and PRONTO-S for these 98 classes with all outcomes represented on the diagonal. 

As both CNNs show 100\% accuracy, for reducing computational complexity we choose PRONTO-S as PRONTO packet detector CNN.

Fig.~\ref{fig:packet-detection-loss} demonstrates training and validation loss over epochs for PRONTO-S architecture. As expected, validation loss closely follows the training loss, which shows the network is not overfitting. The variations in the plot are due to random augmentation of both training and validation batches, which exposes the CNN to different data in every epoch. PRONTO-L loss plot follows the same trend.

\subsection{CFO Estimation Evaluation}\label{sec:eval-cfo}

For the CFO estimation problem, we sort Oracle dataset in ascending order with respect to their CFO labels, $f$s. We pick the first $\sim$50\% portion for training and validation and the second $\sim$50\% for test, so that the training and validation sets contain negative CFOs, whereas the test set contains positive CFOs. In this way, the training and test sets emulate the realistic situation where training set CFO values are in a limited range, but we wish to predict CFOs of other ranges during test, as well. 
For the sake of extensive evaluation, we artificially inject CFOs in the range of [0, $\Delta f/2$] to the test set signals.

\noindent \textbf{CFO Augmentation Impact.}
To show the effect of CFO augmentation step, we first train PRONTO-L with the determined training set without CFO augmentation in the training pipeline. 
We compare predicted CFO values $\hat{Y}^{\CFO}$ at the output of the CNN with true CFO labels $Y^{\CFO}$ and calculate mean absolute error (MAE) for each SNR level (distance) as in (\ref{eq:mae}), where $M$ is the number of test L-LTF signals in each SNR.
\begin{equation}\label{eq:mae}
    \text{MAE} = \frac{1}{M}\sum_{m=0}^{M-1} |Y_m^{\CFO}-\hat{Y}_m^{\CFO}|
\end{equation}

\begin{figure}[t!!!]
\centering
\begin{tikzpicture}
        \begin{axis}[
            axis lines = left,
            xlabel = Average SNR per distance (dB),
            ylabel = {MAE (Hz)},
            ymin=100,
            ymax=10000000,
            ymode = log,
            xmin=7,
            xmax=40,
            height=5.5cm,
            width=0.45\textwidth,
            legend style={at={(1,1)},
            legend cell align={left},
                anchor=north east,legend columns=1},
            ymajorgrids=true,
            ytick = {100,1000,10000},
            yticklabels = {$10^2$,$10^3$,$10^4$},
            grid style={line width=1pt,draw=gray!50},
        ]
        \addplot [thick,mark=o]
        coordinates{ 
        (7.329,25856.9701847)(9.2535,26361.4313517)(13.9017,25979.1706332)(14.5118,25898.2234702)(15.6828,26061.910657)(17.2901,25685.495992)(20.9097,25860.8655753)(22.9087,25511.9232605)(24.7952,25488.2321074)(26.0466,25610.8449265)(39.8717,23550.7463476)
        };
        \addlegendentry{PRONTO-L, w/o CFO aug.}
        \addplot [thick,mark=*]
        coordinates{ 
        (7.329,1001.63763357)(9.2535,647.895769593)(13.9017,315.589494824)(14.5118,287.1767287)(15.6828,257.600452588)(17.2901,222.228883524)(20.9097,160.42218552)(22.9087,138.013828914)(24.7952,120.176216016)(26.0466,116.171830287)(39.8717,100.207625082)
        };
        \addlegendentry{PRONTO-L, w/ CFO aug.}
        \addplot [thick,mark=+]
        coordinates{
        (7.329,1681.38727598)(9.2535,1054.28867281)(13.9017,470.409709029)(14.5118,431.859657306)(15.6828,399.279815429)(17.2901,364.366123119)(20.9097,319.130177057)(22.9087,306.056239133)(24.7952,296.562796516)(26.0466,301.550583378)(39.8717,469.009250123)
        };
        \addlegendentry{PRONTO-S, w/ CFO aug.}
        \end{axis}
    \end{tikzpicture}
    \caption{Mean absolute error (MAE) between coarse CFOs predicted by CNNs using L-LTF signals and by \texttt{MATLAB} using L-STF signals, with and without CFO augmentation.\vspace{-0.3cm}}
    \label{fig:error-mean-absolute}
\end{figure}
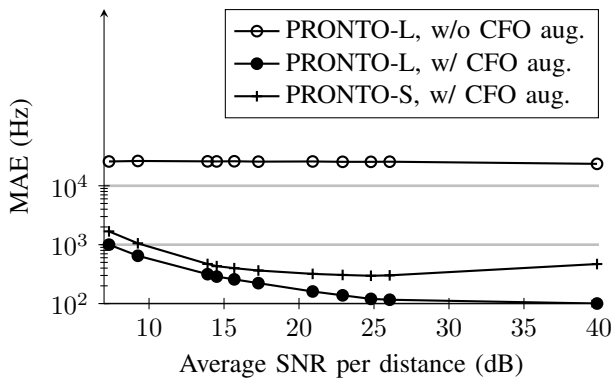

\begin{figure*}[t!!!]
    \centering
    \includegraphics[width=0.9\linewidth]{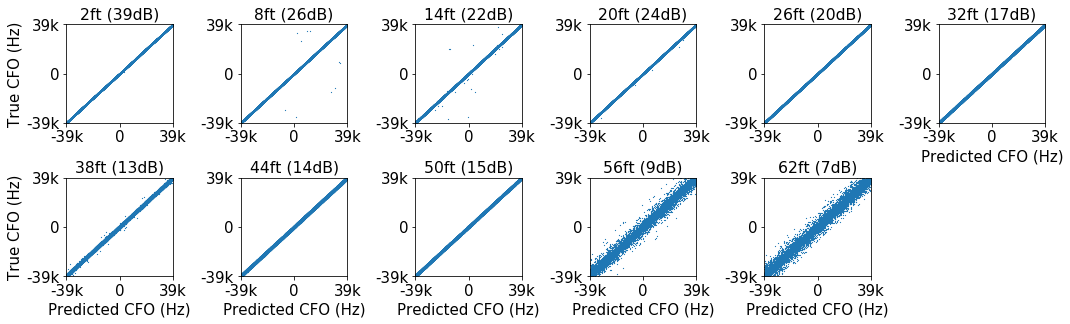}
    \caption{Scatter plots showing the true CFO versus the predicted CFO for PRONTO CFO estimator in 11 different distances (SNRs). The predicted CFO shows a larger divergence from the true CFO in lower SNR regions.}
    \label{fig:result-conf-line}
\end{figure*}

\begin{table*}[h!!!]
    \centering
    \small
    \resizebox{\linewidth}{!}{
    \begin{tabular}{|c|b|d|>{\bfseries}y||b|d|>{\bfseries}y|}
        \hline\hline\xrowht[()]{5pt}
        Distance &\multicolumn{3}{c||}{Traditional receiver (all \texttt{MATLAB})}& \multicolumn{3}{c|}{PRONTO-based receiver} \\
        \cline{2-7}\xrowht[()]{5pt}
        (ft) &\multicolumn{1}{c|}{Coarse CFO (Hz)} & \multicolumn{1}{c|}{Fine CFO (Hz)} & \multicolumn{1}{c||}{{\bf Coarse+Fine (Hz)}} & \multicolumn{1}{c|}{Coarse CFO (PRONTO) (Hz)} & \multicolumn{1}{c|}{Fine CFO (\texttt{MATLAB}) (Hz)} & \multicolumn{1}{c|}{{\bf Coarse+Fine (Hz)}}\\
        \hline
        \xrowht[()]{5pt}
        2 & 31186.47413 & 1.672545869 & 31188.14668 & 31073.47413 $\downarrow$ & 114.6725459 $\uparrow$ & 31188.14668\\
        \hline
        \xrowht[()]{5pt}
        8 & 27935.04853 & -173.4179159 & 27761.63061 & 28056.04853 $\uparrow$ & -294.4179159 $\downarrow$ & 27761.63061\\
        \hline
        \xrowht[()]{5pt}
        14 & 24935.6091 & -24.97061088 & 24910.63849 & 25076.6091 $\uparrow$ & -165.9706109 $\downarrow$ & 24910.63849\\
        \hline
        \xrowht[()]{5pt}
        20 & 22454.2147 & 118.8058465 & 22573.02055 & 22331.2147 $\downarrow$ & 241.8058465 $\uparrow$ & 22573.02055\\
        \hline
        \xrowht[()]{5pt}
        26 & 26613.99321 & -2.242109413 & 26611.7511 & 26773.99321 $\uparrow$ & -162.2421094 $\downarrow$ & 26611.7511\\
        \hline
        \xrowht[()]{5pt}
        32 & 25127.22223 & -153.0018763 & 24974.22035 & 24901.22223 $\downarrow$ & 72.9981237 $\uparrow$ & 24974.22035\\ 
        \hline
        \xrowht[()]{5pt}
        38 & 23959.62959 & 86.35591791 & 24045.9855 & 23641.62959 $\downarrow$ & 404.3559179 $\uparrow$ & 24045.9855\\
        \hline
        \xrowht[()]{5pt}
        44 & 24733.93078 & -196.1821002 & 24537.74868 & 24444.93078 $\downarrow$ & 92.81789983 $\uparrow$ & 24537.74868\\
        \hline
        \xrowht[()]{5pt}
        50 & 24282.74508 & 153.5915397 & 24436.33662 & 24541.74508 $\uparrow$ & -105.4084603 $\downarrow$ & 24436.33662\\
        \hline
        \xrowht[()]{5pt}
        56 & 27072.75937 & -37.8109977 & 27034.94837 & 27739.75937 $\uparrow$ & -704.8109977 $\downarrow$ & 27034.94837\\
        \hline
        \xrowht[()]{5pt}
        62 & 32227.95237 & -5318.056399 & 26909.89597 & 31128.95237 $\downarrow$ & -4219.056399 $\uparrow$ & 26909.89597\\
         \hline\hline
    \end{tabular}
    }
    \caption{Comparison of CFO estimation by traditional and PRONTO-based receivers, evaluated on the OTA WiFi dataset~\cite{oracle-kunal}. We observe that despite the small error in `Coarse CFO (PRONTO)', `Coarse+Fine' (bold column) in both cases of \texttt{MATLAB} and PRONTO lead to the exact same values (given until 5 decimal places). The reason is that \texttt{MATLAB} fine CFO estimator (`Fine CFO (\texttt{MATLAB})') that follows `Coarse CFO (PRONTO)', estimates the residual CFO that remains after the execution of PRONTO, as a part of fine CFO. Hence, no BER degradation occurs in the PRONTO-based receiver in comparison to the traditional receiver.} 
    \label{tab:sum}
\end{table*}

The MAE for the aforementioned case for different SNR levels is shown in Fig.~\ref{fig:error-mean-absolute} under the legend ``PRONTO-L, w/o CFO aug.''.
In a follow-up experiment, we train and test PRONTO-L with the same training and test sets, this time with CFO augmentation in the training pipeline, same as in Fig.~\ref{fig:proposed-cfo}. We calculate the MAE for this case for different SNR levels and show it under the legend ``PRONTO-L w/ CFO aug.'' in Fig.~\ref{fig:error-mean-absolute}. We see that the augmentation block improves the MAE results by up to $\sim$20000 Hz. This large improvement is achieved by exposing the neural network to different random CFOs that are not present in the training set and are artificially injected using the CFO augmentation block during training.

\noindent \textbf{CNN Architecture Impact.} 
In addition to CFO augmentation, the choice of CNN architecture also plays a key role. We train and test PRONTO-S in Fig.~\ref{fig:nn-arch} with CFO augmentation block in the training pipeline. The results for different SNRs are shown as ``PRONTO-S w/ CFO aug.'' in Fig.~\ref{fig:error-mean-absolute}. We observe that the best CFO estimation (lowest MAE) is achieved with a comparatively deeper CNN (PRONTO-L), when CFO augmentation is included in the training pipeline.

\noindent \textbf{Scatter Plots For Coarse CFO Estimation.}
In Fig.~\ref{fig:result-conf-line}, scatter plots of ``True CFO'' versus ``Predicted CFO'' for each distance are demonstrated. The prediction results are achieved using PRONTO-L trained with CFO augmentation block in the training pipeline. We observe that for higher SNRs, we see the presence of a strong diagonal. As the SNR decreases, there is increased deviation from the diagonal line, which shows an increase in error. This is consistent with our observations in Fig.~\ref{fig:error-mean-absolute}.

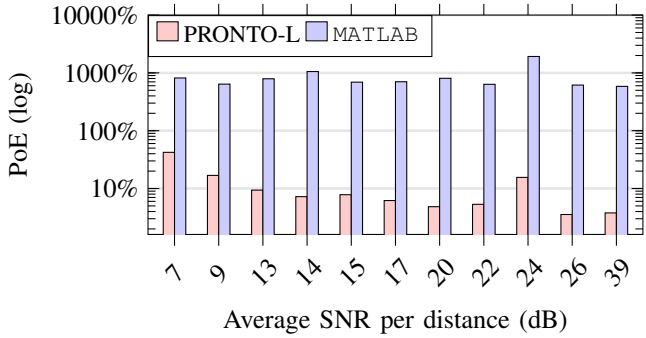
\begin{figure}[t!!!]
\centering
    \begin{tikzpicture}
        \begin{axis}[
            bar width = 0.15cm,
            ybar=0pt,
            height=4.5cm,
            width=0.45\textwidth,
            ymin=-1,
            ymax=10000,
            enlarge x limits=0.06,
            ybar legend,
            legend style={at={(0,1), font=\small}, legend cell align={left},
                anchor=north west,legend columns=2},
            ylabel={PoE (log)},
            xlabel={Average SNR per distance (dB)},
            y label style = {align=center,text width=3cm},
            ymode = log,
            symbolic x coords={1,2,3,4,5,6,7,8,9,10,11},
            xtick=data,
            xticklabels = {7,9,13,14,15,17,20,22,24,26,39},
            x tick label style={align=center, text width=0.3cm, rotate=45}, ytick={1,10,100,1000,10000},
            yticklabels = {0\%,10\%,100\%,1000\%,10000\%},
            ymajorgrids=true,
            grid style={line width=1pt,draw=gray!20},
            ]
        \addplot [black, fill=red!20]
        coordinates{ 
        (1,42.2609002103)(2,16.8613103296)(3,9.38340326971)(4,7.22594104815)(5,7.81413103894)(6,6.19560124087)(7,4.84414063387)(8,5.32952711366)(9,15.5703566878)(10,3.55169555061)(11,3.78442855313)
        };
        \addlegendentry{PRONTO-L}
        \addplot [black, fill=blue!20]
        coordinates{ 
        (1,816.65)(2,639.42)(3,791.06)(4,1055.22)(5,691.63)(6,703.37)(7,806.81)(8,634.8)(9,1923.61)(10,615.77)(11,584.69)
        };
        \addlegendentry{\texttt{MATLAB}}
        \end{axis}
    \end{tikzpicture}
    \caption{Percentage of error (PoE) when \texttt{MATLAB} estimates CFO on coarsely time-synchronized L-LTF signals compared to PRONTO-L estimations.\vspace{-0.5cm}}
    \label{fig:error-percentage}
\end{figure}

\noindent \textbf{Coarse CFO Estimation \texttt{MATLAB} vs. PRONTO.}
To demonstrate PRONTO's ability to estimate coarse CFO compared with traditional methods, we feed the coarsely-time synchronized L-LTF signals in the test set to \texttt{MATLAB} function \texttt{wlanFineCFOEstimate}. This function is designed to estimate CFO using L-LTF signals through the traditional methods, and is used in fine frequency synchronization in the traditional receiver. 

We collect predictions on test set signals from PRONTO-L and then calculate the percentage of error (PoE) in each SNR (distance) for both \texttt{MATLAB} calculations and PRONTO-L predictions, using (\ref{eq:percentage-of-error}).\vspace{-0.3cm}

\begin{equation}\label{eq:percentage-of-error}
    \text{PoE} = \frac{1}{M}\sum_{m=0}^{M-1} \frac{|Y_m^{\CFO}-\hat{Y}_m^{\CFO}|}{|Y_m^{\CFO}|}
\end{equation}

As demonstrated in Fig.~\ref{fig:error-percentage} the \texttt{MATLAB} function yields very large errors of upto $\sim$2000\% when fed with coarsely time-synchronized L-LTFs, as it fails to estimate the coarse CFO on L-LTF signals. In contrast, PRONTO-L can estimate coarse CFO on L-LTF signals with errors as small as 3\%.


\noindent \textbf{Coarse CFO Estimation Error Impact on BER.}
In Fig.~\ref{fig:error-mean-absolute}, we demonstrate that our CFO estimator CNN (without L-STF) performs within an error of $\sim$300 Hz averaged over all SNRs (distances) in reference to the traditional coarse CFO estimator (that uses L-STF). The question is: \emph{How much BER degradation is caused by this small error in PRONTO's coarse CFO estimator?}

To answer this question, we measure the BER over the dataset when the receiver works with the legacy coarse CFO estimator, as well as with the coarse CFO estimator CNN, PRONTO-L. We observe that the two BERs are exactly the same. Therefore, the answer to the above question is \emph{PRONTO causes no BER degradation}. 

We explain below as to why there is no BER degradation, despite some error in coarse CFO estimation. The fine CFO estimation function, which is implemented through the legacy approach is shown in Fig.~\ref{fig:overview}(b). This function tunes the fine CFO to compensate for the small errors in coarse CFO. Table~\ref{tab:sum} shows 11 test set packets with randomly injected CFOs for each distance from 2 ft to 62 ft. We observe that the traditional receiver (columns 2,3,4) estimates coarse and fine CFOs for each packet that leads to a total CFO (Coarse+Fine). In the PRONTO-based receiver (columns 5,6,7), the up/down arrows under the Coarse CFO (PRONTO) (column 5) show that the coarse CFO estimated by PRONTO increases/decreases in reference to coarse CFO of the traditional receiver (column 2). However, the fine CFO estimator that comes after coarse CFO estimation by PRONTO (column 6) compensates in the opposite direction. 
In this way, the total CFO (Coarse+Fine) for the traditional receiver (column 4) and PRONTO-based receiver (column 7) are calculated to be exactly equal to each other.

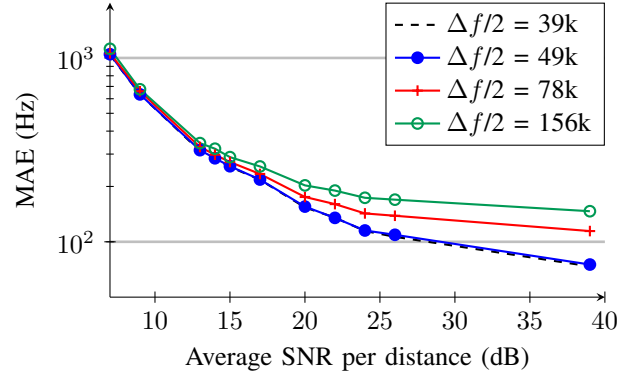
\begin{figure}[t!!!]
\centering
\begin{tikzpicture}
        \begin{axis}[
            axis lines = left,
            xlabel = Average SNR per distance (dB),
            ylabel = {MAE (Hz)},
            ymin=50,
            ymax=2000,
            ymode = log,
            xmin=7,
            xmax=40,
            height=5.5cm,
            width=0.45\textwidth,
            legend style={at={(1,1)},
            legend cell align={left},
                anchor=north east,legend columns=1},
            ymajorgrids=true,
            ytick = {100,1000},
            yticklabels = {$10^2$,$10^3$},
            grid style={line width=1pt,draw=gray!50},
        ]
        \addplot [thick,dashed]
        coordinates{ 
        (7,1040.8709692954583)(9,633.5741367303224)(13,311.8465412609886)(14,284.1578730050179)(15,255.6277072493824)(17,217.57728934668847)(20,153.86589008597554)(22,135.2971007129926)(24,114.52746104555105)(26,106.46152970501258)(39,73.55237031320468)
        };
        \addlegendentry{$\Delta f$/2 = 39k}
        \addplot [thick,blue,mark=*]
        coordinates{ 
        (7,1047.6936578839536)(9,633.6329674469453)(13,315.1007146256475)(14,284.6721614513469)(15,257.63472086477805)(17,218.11464699081694)(20,155.45490276288623)(22,134.9423022824833)(24,115.22497392864068)(26,109.17295777173521)(39,75.2210483010342)
        };
        \addlegendentry{$\Delta f$/2 = 49k}
        \addplot [thick,red,mark=+]
        coordinates{(7,1059.8941878094458)(9,658.9282085060433)(13,328.00196306918497)(14,297.6520779600129)(15,271.568203722749)(17,233.89517683825446)(20,175.32138611231585)(22,160.34970946390823)(24,142.68157553140162)(26,138.554116908219)(39,114.39169979342017)
        };
        \addlegendentry{$\Delta f$/2 = 78k}
        \addplot [thick,ForestGreen,mark=o]
        coordinates{
        (7,1121.3860150285707)(9,675.3341766304356)(13,344.22702914388833)(14,320.9202087851151)(15,288.7792486556604)(17,256.926954194737)(20,202.7362425843567)(22,189.943070272511)(24,173.5367219115809)(26,169.52634159016273)(39,146.6768663481802)
        };
        \addlegendentry{$\Delta f$/2 = 156k}
        \end{axis}
    \end{tikzpicture}
    \caption{Mean absolute error (MAE) between coarse CFOs predicted by CNNs and the imposed CFO when different values are chosen for $\Delta f$ of CFO augmentation block.\vspace{-0.3cm}}
    \label{fig:delta-f}
\end{figure}

\noindent \textbf{Coarse CFO Values Beyond Sub-carrier Spacing.} The \texttt{MATLAB} function for coarse CFO estimation (i.e., \texttt{wlanCoarseCFOEstimate}~\cite{wlanCoarseCFOEstimate}) can estimate coarse CFOs as large as twice the sub-carrier spacing. To show that PRONTO-L CFO estimator can perform similarly, we draw $f_{\text{off.}}$ from $U(-\frac{\Delta f}{2} , \frac{\Delta f}{2})$ as explained in Section~\ref{sec:proposed-cfo}, and vary the ranges as [-39k, 39k] (same $\Delta f$ as shown in Table~\ref{tab:parameters}), [-49k, 49k], [-78k, 78k], and [-156k, 156k]. For each range, we train and test PRONTO-L CFO estimator with $\Delta f$s from the specific range and show MAE versus SNR in Fig.~\ref{fig:delta-f}. We observe that MAE in higher SNRs slightly increases from 73 Hz (39k graph) to 146 Hz (156k graph), however, it remains smaller than the largest MAE for $\Delta f/2$=39k. From Table~\ref{tab:sum}, this CFO MAE is compensated by fine CFO estimation, and does not impact the BER. Therefore, the PRONTO-L is able to perform coarse CFO estimation of CFOs as large as twice the sub-carrier spacing similar to the \texttt{MATLAB} function.

\subsection{PRONTO Performance in a New Environment} \label{sec:eval-environment}

To evaluate packet detection in a new environment, we test the PRONTO-S architecture, pretrained on Oracle dataset in Section~\ref{sec:eval-packet}, and we test it on packets from Arena dataset. The reason this test is possible is that although sampling frequency is different in Oracle (training) and Arena (test) datasets, the L-LTF length is equal to 160 I/Q samples for sampling frequencies between 5 to 20 MHz. Therefore, the number of classes for packet detection in Arena dataset is the same as Oracle dataset as shown in Fig.~\ref{fig:proposed-classes}. We observe the same 100\% classification accuracy for the Arena dataset without re-training the neural network.

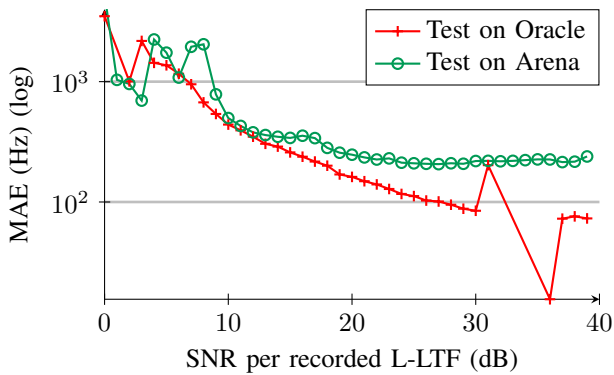
\begin{figure}[t!!!]
\centering
\begin{tikzpicture}
        \begin{axis}[
            axis lines = left,
            xlabel = SNR per recorded L-LTF (dB),
            ylabel = {MAE (Hz) (log)},
            ymin=0,
            ymax=4000,
            ymode = log,
            xmin=0,
            xmax=40,
            height=0.3\textwidth,
            width=0.45\textwidth,
            xtick={0,10,20,30,40},
            legend style={at={(1,1)},
            legend cell align={left},
                anchor=north east,legend columns=1},
            ymajorgrids=true,
            ytick = {1,100,1000},
            yticklabels = {1,$10^2$,$10^3$},
            grid style={line width=1pt,draw=gray!50},
        ]
        \addplot [thick,red,mark=+]
        coordinates{
        (0,3497.6184214022724)(2,1005.9484314018357)(3,2179.3080276847645)(4,1432.851969525952)(5,1367.553472419986)(6,1164.1403094367258)(7,951.0419248401894)(8,672.6265445885933)(9,538.918362355146)(10,439.88119112118636)(11,393.9099536135914)(12,348.85885025228197)(13,304.4343452727607)(14,288.6188731635824)(15,258.29885927504716)(16,237.80383949029343)(17,216.78055438401645)(18,200.25187946634654)(19,169.39826484373745)(20,161.647501270593)(21,148.8416227465067)(22,139.94691804044106)(23,128.26189122743992)(24,116.62974289128508)(25,111.95033591359862)(26,102.82043133274227)(27,100.98712438970755)(28,94.91925868455712)(29,87.92227692324359)(30,84.55745864016444)(31,202.4663896846323)(36,15.554465882342129)(37,72.67531849551335)(38,75.97164253047684)(39,73.03883221157288)
        };
        \addlegendentry{Test on Oracle}
        \addplot [thick,ForestGreen,mark=o]
        coordinates{
        (0,5312.880283073027)(1,1034.85714638247)(2,956.8738915695955)(3,696.0793325355517)(4,2240.740371234793)(5,1741.2428300857046)(6,1081.2613245083057)(7,1948.9149544955537)(8,2043.3502470958429)(9,784.3663906001391)(10,498.2215275297874)(11,428.5118513484339)(12,379.04357845988795)(13,359.9524511791578)(14,348.1736452482007)(15,340.78768910682965)(16,355.43242850075325)(17,339.4658797664709)(18,281.6782688291015)(19,256.7997463370702)(20,247.1582538131708)(21,234.70035815025437)(22,225.41926804492516)(23,230.3300872875443)(24,212.1329618158563)(25,209.6518310906551)(26,207.65443199746377)(27,206.31675692618512)(28,209.80777574884067)(29,207.76786486026637)(30,218.95284019126757)(31,219.69727130077192)(32,217.84569588705554)(33,219.6690260987224)(34,222.730198507879)(35,225.96237697768703)(36,225.62354459180287)(37,214.506949927877)(38,216.30324657915077)(39,238.93441069133405)
        };
        \addlegendentry{Test on Arena}
        \end{axis}
    \end{tikzpicture}
    \caption{Mean Absolute Error (MAE) for CFO estimation in test sets of Oracle and Arena (two different environments). For both tests, the same model trained on data from Oracle environment is used.}\vspace{-0.3cm}
    \label{fig:arena}
\end{figure}

To evaluate CFO estimation in a new environment, we train two PRONTO-L CNNs on the training portion of the Oracle dataset, one with $\Delta f$=78 kHz ($f_{\text{off.}}$$\sim$$U$(-39k, 39k)), and another with $\Delta f$=156 kHz ($f_{\text{off.}}$$\sim$$U$(-78k, 78k)). In this way, we set $\Delta f$ for each CNN equal to the sub-carrier spacing in each of the 5 MHz and 10 MHz Oracle and Arena datasets, respectively. However, in both cases, we set $f_s$ as the sampling rate of the training dataset, i.e., 5 MHz, and we train both CNNs on Oracle dataset. Since there is no distance concept in the Arena dataset, for a point to point comparison, we calculate the per-L-LTF frequency domain SNR using \texttt{MATLAB} function \texttt{helperNoiseEstimate}, as explained in Section~\ref{sec:setup-extraction}, without averaging over distances. We test the trained models on their respective test datasets from Oracle (same environment as the training dataset) and Arena (different environment from the training dataset), and measure the MAE at each SNR for each test set. 
The average of MAEs shown in Fig.~\ref{fig:arena} are 247 Hz and 270 Hz for Oracle and Arena datasets, respectively. We observe that with changing the environment, coarse CFO error increases by 9\%, however, the MAE is still in the range of Table~\ref{tab:sum}, and hence the residual CFO is compensated by fine CFO estimation process. Therefore, no compromise in BER happens if PRONTO is trained in an indoor environment, and is deployed in a new indoor environment.


\subsection{Computation Cost}\label{sec:eval-computation}

We evaluate PRONTO in terms of computation cost by counting
floating point operations (FLOPs), similar to the state-of-the-art (SOTA)~\cite{packet-detect-cfo,packet-detection,coarse-cfo,tong-edge}, and run-time on 4 different commonly used platforms of desktop and edge CPUs and GPUs, as shown in Table~\ref{tab:computation} and listed below: 

\begin{enumerate}
    \item Desktop-CPU: AMD Ryzen Threadripper 2950X 16-Core Processor
    \item Desktop-GPU: NVIDIA GeForce RTX 2080 GPU
    \item Jetson TX2-CPU: Quad-Core ARM Cortex-A57
    \item Jetson TX2-GPU: NVIDIA Pascal GPU
\end{enumerate}
    
As explained in Sections~\ref{sec:eval-packet} and~\ref{sec:eval-cfo}, in the PRONTO-based wireless receiver we use PRONTO-S architecture for packet detection and PRONTO-L for coarse CFO estimation (shown in Fig.~\ref{fig:nn-arch}).

\begin{table*}[h!!!]
    \centering
    \small
    \resizebox{0.85\linewidth}{!}{
    \begin{tabular}{c|l|l||c|c||c|c|}
        \cline{2-7}\xrowht[()]{7pt}
        &\multicolumn{2}{c||}{Task} 
        &\multicolumn{2}{c||}{Packet Detection}
        &\multicolumn{2}{c|}{Coarse CFO Estimation} \\
        \cline{2-7}
        \xrowht[()]{7pt}
        &\multicolumn{2}{c||}{Method} & \texttt{MATLAB} & PRONTO-S & \texttt{MATLAB} & PRONTO-L \\
        \cline{2-7} 
        \noalign{\vskip\doublerulesep
         \vskip-\arrayrulewidth}
        \cline{4-7}
        \xrowht[()]{7pt}
        &\multicolumn{2}{c||}{Computational complexity (FLOPs)} & 2400 & 1.7M & 417 & 40.8M \\
        \cline{2-7}
        \xrowht[()]{7pt}
        &Run Time& Desktop-CPU & \textbf{30.8$\pm$0.62} & 1.09$\pm$0.12 & \textbf{0.05$\pm$0.02} & 3.15$\pm$0.29 \\
        \cline{3-7}
        \xrowht[()]{7pt}
        &(ms) & Desktop-GPU & - & 1.9$\pm$0.19 & - & 1.88$\pm$0.22 \\
        \cline{3-7}
        \xrowht[()]{7pt}
        &Batch size=1 & Jetson TX2-CPU & - & 4.91$\pm$0.54 & - & 9.88$\pm$0.38 \\
        \cline{3-7}
        \xrowht[()]{7pt}
        && Jetson TX2-GPU & - & 8.98$\pm$0.61 & - & 7.99$\pm$1.01 \\
        \cline{2-7}
        \xrowht[()]{7pt}
        &Run Time & \multirow{2}{*}{Desktop-GPU} & - & 2.02$\pm$0.16  & - & 3.34$\pm$0.27 \\
        &(ms) &&&\textbf{(0.015 ms per input)}&&\textbf{(0.026 ms per input)} \\
        \cline{3-7}
        \xrowht[()]{7pt}
        & Batch size=128 &\multirow{2}{*}{Jetson TX2-GPU} & - & 17.63$\pm$0.89 & - & 53.19$\pm$1.64 \\
        &&&&\textbf{(0.137 ms per input)}&&\textbf{(0.415 ms per input)} \\
        \cline{2-7}
    \end{tabular}
    }
    \caption{Comparison of FLOPs and run-time in traditional (\texttt{MATLAB}) and PRONTO (CNN) algorithms for packet detection and coarse CFO estimation. Run-durations are measured 100 times on each of the 4 different platforms, and results are demonstrated as \emph{mean$\pm$standard deviation} of these 100 time durations.}
    \label{tab:computation}
\end{table*}

\noindent \textbf{FLOPs and CPU Run-Time.} We count FLOPs in both \texttt{MATLAB} functions~\cite{flops-matlab} and PRONTO CNNs for the two tasks of packet detection and coarse CFO estimation, and show them in the third row of Table~\ref{tab:computation}. As expected, PRONTO-L that is a larger CNN consists of more FLOPs compared to PRONTO-S (40.8M FLOPs for PRONTO-L versus 1.7M FLOPs for PRONTO-S). To have a fair comparison between traditional (\texttt{MATLAB}) and PRONTO algorithms run-time, we first run all the algorithms with \emph{one} input (batch size=1) on the same platform of Desktop-CPU. To account for operating system interrupts and run-time fluctuations, we run \texttt{MATLAB} implementations for each algorithm of packet detection and coarse CFO estimation 100 times in a loop and measure 100 individual run-time durations. We calculate mean and standard deviation of these run-times and show them in the fourth row of Table~\ref{tab:computation} as \emph{mean}$\pm$\emph{standard deviation}. We repeat the same process of measuring run-time on CPU for PRONTO-S and PRONTO-L CNNs with batch size 1. We observe that PRONTO CNNs consist of much larger number of FLOPs compared to their traditional \texttt{MATLAB} counterparts. However, \texttt{MATLAB} packet detector function takes longer to run compared to PRONTO-S. The reason for this is that in the traditional \texttt{MATLAB} packet detection algorithm there are many comparison statements and branches that contribute to increasing the time, but are not counted as FLOPs.

\noindent \textbf{CPU/GPU Comparison.} We run PRONTO-S and PRONTO-L on Desktop-GPU and measure run-time for one input L-LTF (batch size=1) during the same process as CPU run-time measurements. We observe that in the case of PRONTO-L GPU implementation decreases CPU mean run-time by 40\% (1.8 ms for GPU compared to 3.15 ms for CPU). However, in the case of PRONTO-S GPU implementation increases CPU mean run-time by 65\% (1.9 ms for GPU versus 1.09 ms for CPU). The reason for this is that GPU hardware consists of a large number of processing cores that run with a slower clock compared to CPU clock. GPU implementation is only beneficial when the algorithms are realized with compute-intensive large matrix multiplications, which is the case for large neural networks. Smaller CNNs specially if run with only one input have limited opportunity for GPU parallelization compared to larger CNNs. In other words, in the case of smaller CNN (PRONTO-S) GPU resources are under-utilized, which leads to longer run-time compared to CPU implementation.
For the same reason, PRONTO-S run-time on GPU is larger than that of PRONTO-L despite PRONTO-S being smaller than PRONTO-L.

\begin{figure}
    \centering
    \includegraphics[width=0.25\textwidth]{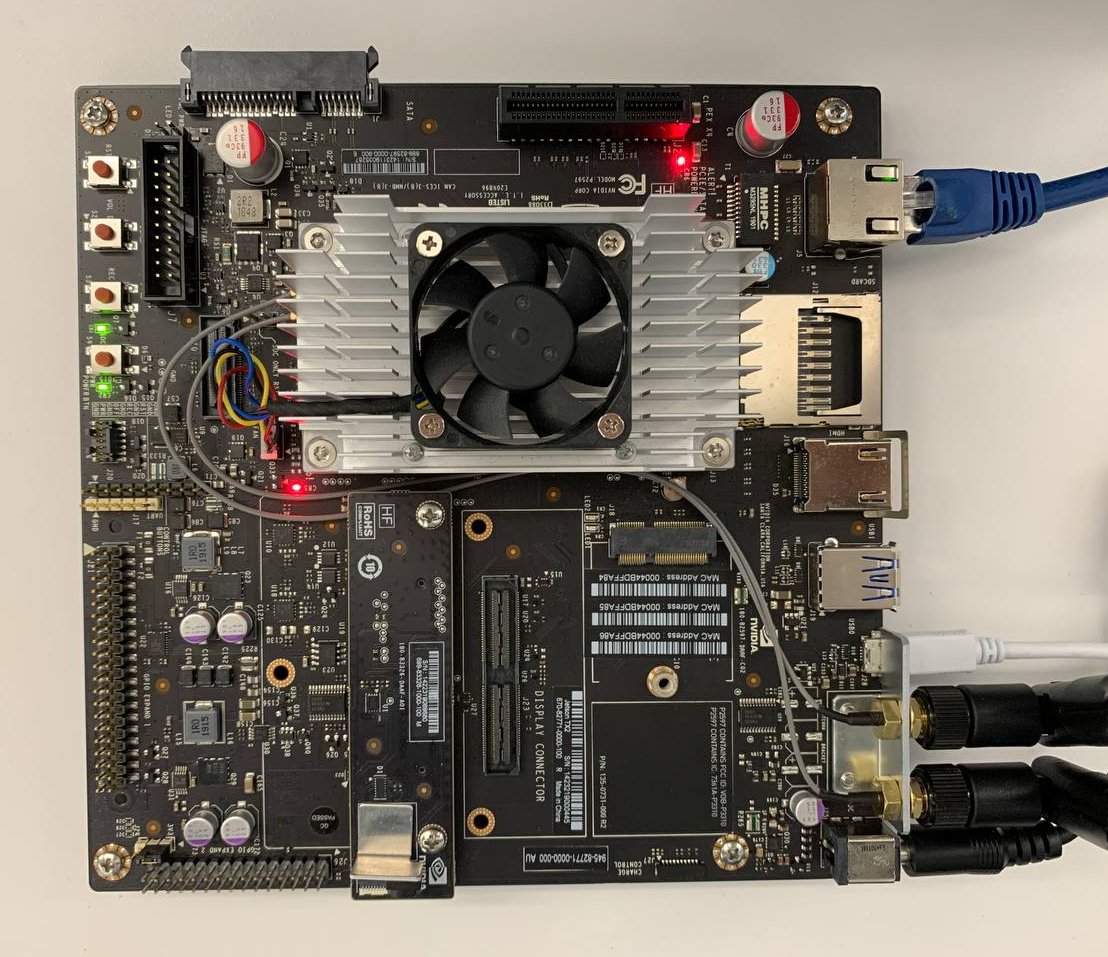}
    \caption{NVIDIA Jetson TX2 used for edge implementation.}
    \label{fig:tx2}
\end{figure}

\noindent \textbf{Edge Implementation.} In order to measure PRONTO run time on a power-efficient edge device, we run PRONTO-S and PRONTO-L on NVIDIA Jetson TX2, an embedded AI computing device shown in Fig.~\ref{fig:tx2}. We can access an ARM Cortex-A57 processor as well as NVIDIA Pascal GPU on the TX2. We run one input L-LTF (batch size=1) through PRONTO-S and PRONTO-L on the ARM core and the GPU of the TX2 and measure mean and standard deviation of the run-time same as before. In Table~\ref{tab:computation} we observe that TX2-GPU decreases PRONTO-L run-time but increases PRONTO-S run-time compared to the run-time on the TX2-CPU core (for the same reason explained earlier). As expected, the run-time of PRONTO on the TX2 is overall larger compared to the powerful desktop computer.

\noindent \textbf{Large Batch Size Effect.} We explore the effect of feeding in large batch size of input (batch size=128) on effective GPU parallelization and PRONTO run-time. We repeat the run-time measuring procedure in a loop that runs 100 times on Desktop-GPU and TX2-GPU, however, instead of just one input (batch size= 1) we feed in 128 inputs together (batch size= 128). These 128 inputs are parallelized on the GPU architecture instead of running sequentially. We observe that the average run-time for PRONTO-S and PRONTO-L equals 2.02 ms and 3.34 ms, respectively, on the Desktop-GPU. If the run-time periods are averaged over the number of inputs, we achieve 0.015 ms and 0.026 ms per input for PRONTO-S and PRONTO-L, respectively. 
Similarly, on TX2-GPU run-time per input decreases to 0.137 ms and 0.415 ms for PRONTO-S and PRONTO-L, respectively, compared to batch size 1 on the same platform. 

\noindent \textbf{Run-time Comparison with Commercial WiFi Card.} For an example WiFi card with 5 MHz sampling rate, each L-LTF is received in 32 $\mu s$. Therefore, each of packet detection and coarse CFO estimation tasks must happen within 32 $\mu s$ for the card to be able to receive streaming signals without packet loss. The reported GPU run-time for PRONTO (15 $\mu s$ and 26 $\mu s$) shows promising results wherein with a custom chip design, a fully-pipelined WiFi card receiver can leverage PRONTO for packet synchronization without packet loss in higher sampling rates. As an example, the decoder neural network proposed in~\cite{nasim-spinn} with $4.45\times10^{10}$ FLOPs that is bigger than both PRONTO-S and PRONTO-L (with $1.7\times10^6$ and $40.8\times10^6$ FLOPs, respectively) works in 100 MHz frequency on FPGAs. A custom designed chip can boost the working frequency of these NNs even further and provide PRONTO timing efficiency in high sampling rate WiFi protocols.


\subsection{Comparison with State-of-the-art (SOTA)}
\label{sec:comparison}

\begin{table}[t!!!]
    \centering
    \small
    \resizebox{0.5\textwidth}{!}{%
    \begin{tabular}{lcccc}
        \hline
        \xrowht[()]{7pt}
        & Ninkovic \textsl{et al.} & Ninkovic \textsl{et al.} & Zhou \textsl{et al.} & \textbf{PRONTO} \\
        & \cite{packet-detection} & \cite{packet-detect-cfo} & \cite{coarse-cfo} & (this paper) \\
        \hline
        \xrowht[()]{7pt}
        Performed packet detection & \Checkmark & \Checkmark & - & \textbf{\Checkmark}\\
        \xrowht[()]{7pt}
        Performed CFO estimation & - & \Checkmark & \Checkmark & \Checkmark\\
        \xrowht[()]{7pt}
        Preamble field used & L-STF & L-STF & L-STF & \textbf{L-LTF} \\
        \xrowht[()]{7pt}
        Used over the air data & - & \Checkmark & - & \textbf{\Checkmark} \\
        \xrowht[()]{7pt}
        Studied BER degradation & - & - & - & \textbf{\Checkmark} \\
        \xrowht[()]{7pt}
        Bandwidth (MHz) & 1 & 1 & 10 & \textbf{5} \\
        \xrowht[()]{7pt}
        Packet detection FLOPs & 200M & 200M & - & \textbf{1.7M} \\
        \xrowht[()]{7pt}
        CFO estimation FLOPs & $>$8738 & - & - & \textbf{40.8M} \\
        \xrowht[()]{7pt}
        Edge implementation & - & - & - & \textbf{\Checkmark} \\
        \xrowht[()]{7pt}
        Packet detection miss rate & 0.7\% & 0.4\% & - & \textbf{0\%} \\
        \xrowht[()]{7pt}
        Preamble length reduction & 0\% & 0\% & 0\% & \textbf{upto 40\%} \\
        \hline
    \end{tabular}
    }
    \caption{Comparing PRONTO with the state-of-the-art.}
    \label{tab:comparison}
\end{table}

Finally, we demonstrate the superiority of PRONTO over the SOTA techniques by evaluating its performance on various metrics as presented in Table~\ref{tab:comparison}. As PRONTO is a first-of-its-kind that uses only L-LTF for both packet detection and coarse CFO estimation, we do not have a direct comparison point among the related work. However, we compare our results with the SOTA approaches which use deep learning for packet detection and coarse CFO estimation using the traditional inputs for these tasks (i.e., the L-STF signals). 
In the following, we list the SOTA-related work that we use for comparison which were also previously discussed in Section~\ref{sec:intro}.

\begin{enumerate}
    \item Ninkovic~\textsl{et al.}~\cite{packet-detection}: Proposes deep learning for estimating the packet (L-STF) start index as a regression problem on simulated data. 
    \item Ninkovic~\textsl{et al.}~\cite{packet-detect-cfo}: Proposes deep learning for estimating the packet (L-STF) start index as well as CFO, both as regression problems, on OTA data.
    \item Zhou~\textsl{et al.}~\cite{coarse-cfo}: Proposes deep learning for detecting the class of CFO with L-STF of simulated data as a classification problem.
\end{enumerate}

Our comparison outcomes are shown in Table~\ref{tab:comparison} and the highlights are described in the following. 

\noindent \textbf{Novelty and Superiority.} Our novelty of \emph{not} using L-STF for packet detection and coarse CFO estimation provides the possibility of removing the L-STF, which reduces WiFi preamble length by upto 40\%.

\noindent \textbf{Model Size and Complexity.}
We show that our packet detector CNN (PRONTO-S) is much lighter compared to the architectures used in the SOTA. We show 1.7M FLOPs for packet detection compared to 200M FLOPs presented in the SOTA~\cite{packet-detect-cfo}.

\noindent \textbf{Communication Quality.} While we study communication quality (i.e., BER) in our proposed PRONTO-based receiver, none of the studied related work consider the effect of errors in their packet detection and coarse CFO estimation on the BER. 

\noindent \textbf{Edge Implementation.} None of the studied related work present edge implementation and run-time results for deployment of packet detection and CFO estimation neural networks in wireless receivers. However, we present edge implementation results on different platforms for conventional blocks as well as PRONTO.

\noindent \textbf{Packet Detection Miss Rate.} From another perspective, \cite{packet-detection} and \cite{packet-detect-cfo} report 0.7\% and 0.3\% miss detection rate, respectively, for packet detection with input size 160 (same input size as PRONTO). However, our 100\% packet detection accuracy shows that all the packets detected by the traditional \texttt{MATLAB}-based packet detection algorithm are detected by PRONTO too and no packets are missed (0\% miss rate). We believe that our robust packet detection algorithm is achieved by two techniques. Firstly, we treat packet detection as a classification problem instead of the less accurate regression proposed by the SOTA~\cite{packet-detect-cfo,packet-detection}. Secondly, we use a data augmentation block that emulates different shifts in the L-LTF, accompanied by periods of noise, that contribute to training a more robust packet detection CNN. 

To the best of our knowledge, there is no related work on packet detection and CFO estimation using L-LTF signals to be used as a direct comparison point. However, we establish the efficacy of PRONTO by achieving same BER as the traditional method after eliminating upto 40\% of the preamble.

\section{Conclusions}\label{sec:conclusion}
PRONTO reduces the packet preamble overhead by eliminating the need for the first field of the preamble (i.e., L-STF), while remaining compliant with legacy standard waveforms that contain the entire preamble. To decode packets without L-STF, PRONTO performs coarse time and frequency synchronizations using the next field of the preamble (i.e., the L-LTF). PRONTO consists of two CNN-based modules for packet detection (coarse time synchronization) and coarse CFO estimation, along with their corresponding augmentation steps for training robust CNNs. We evaluated PRONTO on two OTA WiFi datasets collected from a testbed of SDRs and showed that packet detector CNN yields 100\% accuracy with a modest error as small as 3\% in coarse CFO estimation. We demonstrate that this small error in coarse CFO estimation does not degrade the BER. Finally, by comparing to neural networks implemented on custom hardware in the existing literature, we envision that with properly pipelined customized hardware, PRONTO can keep up with WiFi transmissions with high sampling rates.

\section*{Acknowledgement}
This work is supported by the US National Science Foundation (NSF) CNS-1923789 award.

\bibliographystyle{ieeetr} 

\bibliography{ref} 

\begin{IEEEbiography}
[{\includegraphics[width=1in,height=1.5in,clip,keepaspectratio]{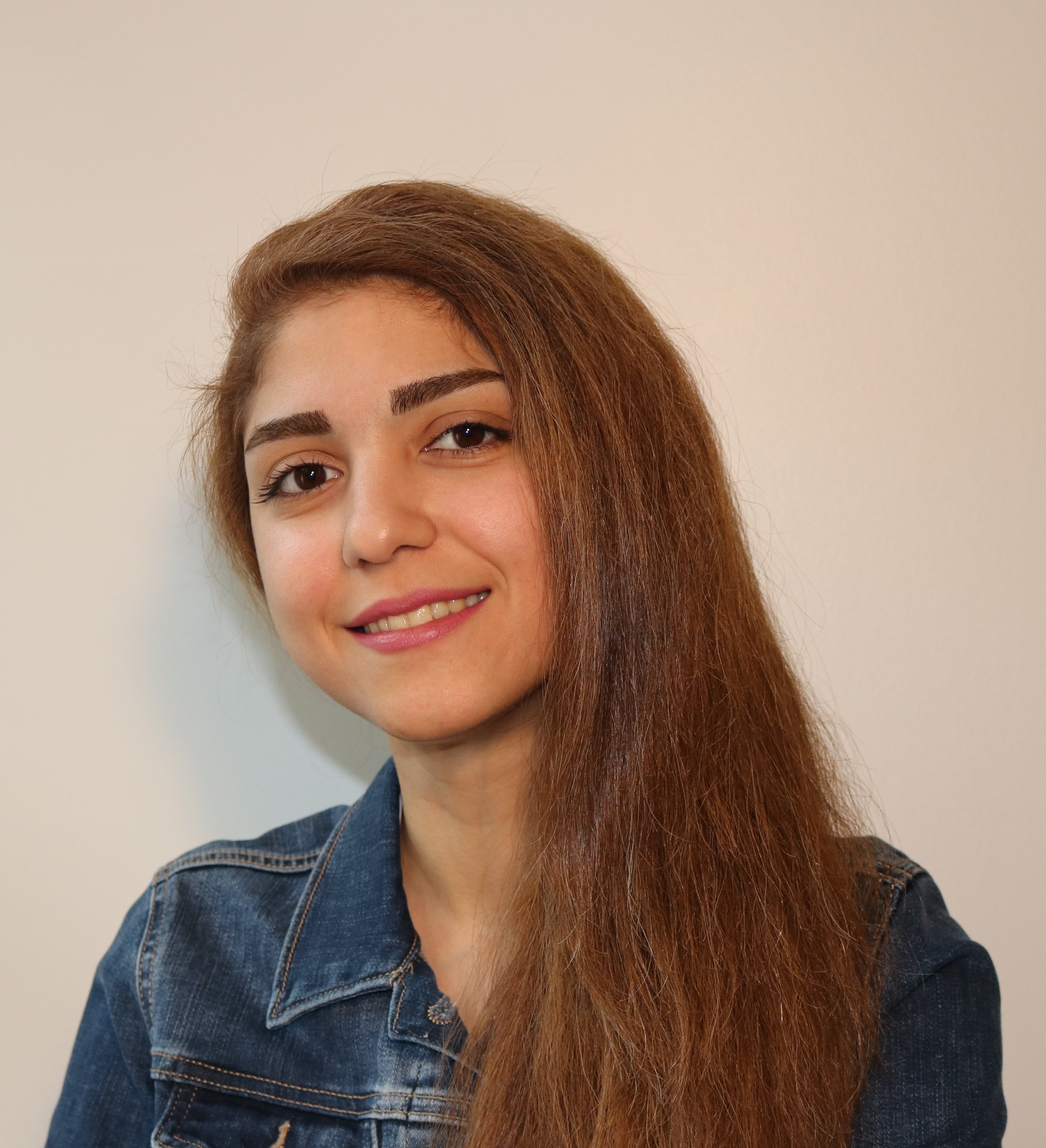}}] 
{Nasim Soltani} is a PhD candidate at the Department of Electrical and Computer Engineering, Northeastern University, Boston, MA. She started her PhD under supervision of Professor Kaushik Chowdhury in 2018. Her research area is ML applications for wireless systems. Her interests are within the physical layer of WiFi and cellular systems, specifically RF fingerprinting, neural network-based receiver design, signal classification, and secure communication.
\end{IEEEbiography}
\vskip -2\baselineskip plus -1fil

\begin{IEEEbiography}
[{\includegraphics[width=1in,height=1.5in,clip,keepaspectratio]{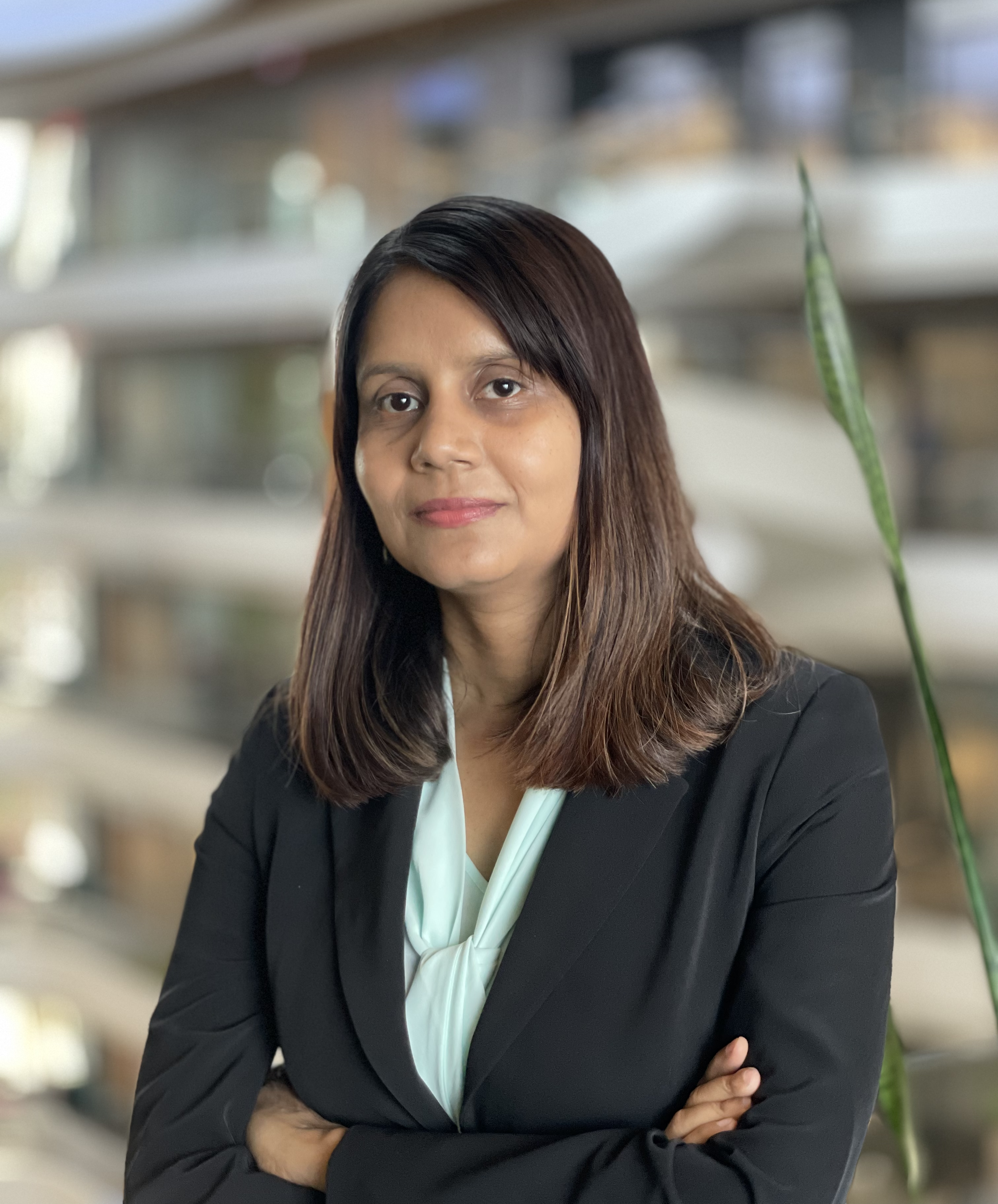}}] 
{Debashri Roy} is currently an Associate Research Scientist at the Department of Electrical and Computer Engineering, Northeastern University. She received her MS (2018) and PhD (2020) degrees in Computer Science from University of Central Florida, USA. She was an experiential AI postdoctoral fellow at Northeastern University (2020-2021). Her research interests are in the areas of AI/ML enabled technologies in wireless communication, multimodal data fusion, network orchestration and nextG networks. 
\end{IEEEbiography}
\vskip -2\baselineskip plus -1fil

\begin{IEEEbiography}
[{\includegraphics[width=1in,height=1.25in,clip,keepaspectratio]{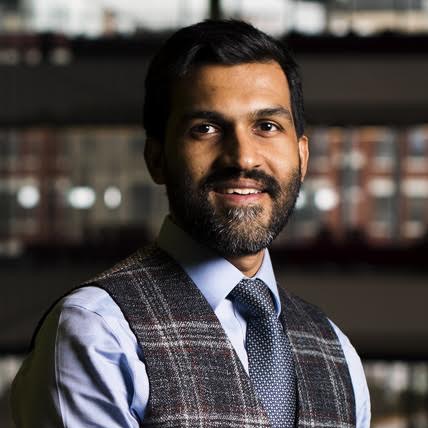}}]{Kaushik Chowdhury} is a Professor at Northeastern University, Boston, MA. 
He is presently a co-director of the Platforms for Advanced Wireless Research (PAWR) project office. His current research interests involve systems aspects of networked robotics, machine learning for agile spectrum sensing/access, wireless energy transfer, and large-scale experimental deployment of emerging wireless technologies.
\end{IEEEbiography}

\end{document}